%% file: neus2025-sample.tex
\documentclass[final]{neus2025}

\input{preamble}
\input{math_commands}

\title[Span-Expanded Attention]{Expansion Span: Combining Fading Memory and Retrieval in Hybrid State Space Models}

\coltauthor{\Name{Elvis Nunez} \Email{elvisnun@amazon.com}\\
 \Name{Luca Zancato} \Email{zancato@amazon.com}\\
 \Name{Benjamin Bowman} \Email{bowmaben@amazon.com}\\
 \Name{Aditya Golatkar} \Email{agolatka@amazon.com}\\
 \Name{Wei Xia} \Email{wxia@amazon.com}\\
 \Name{Stefano Soatto} \Email{soattos@amazon.com}\\
 \addr AWS AI Labs}

\begin{document}

\maketitle


\input{sections/abstract}    
\input{sections/intro}

\input{sections/background}
\input{sections/method}
\input{sections/experiments_no_llama}

\input{sections/conclusion}


\bibliography{latex-template/bibliography}

\appendix
\input{sections/appendix_llama_overflow}

\end{document}

%% file: preamble.tex
\usepackage{times}

\usepackage{microtype}
\usepackage{booktabs} 

\usepackage{mathtools}

\usepackage[capitalize,noabbrev]{cleveref}

\usepackage{caption}
\usepackage{wrapfig}

\usepackage{subcaption}
\usepackage{array}
\usepackage[utf8]{inputenc} 
\usepackage{multirow}
\usepackage{colortbl}
\usepackage{adjustbox}
\usepackage{soul}
\usepackage{makecell}

\usepackage{float}

\newcommand{\ourlora}{HyLoRA}
\newcommand{\ourlorasp}{HyLoRA }
\newcommand{\ourattndef}{Span-Expanded Attention}
\newcommand{\ourattndefsp}{Span-Expanded Attention }
\newcommand{\ourattn}{SE-Attn}
\newcommand{\ourattnsp}{SE-Attn }

\newcommand{\fullattn}{Full-Attn}
\newcommand{\fullattnsp}{Full-Attn }

\newcommand{\swattn}{SW-Attn}
\newcommand{\swattnsp}{SW-Attn }

\newcommand{\ssattn}{$S^2$-Attn}
\newcommand{\ssattnsp}{$S^2$-Attn }

\newcommand{\topk}{k }

%% file: math_commands.tex
\newcommand{\seqlen}{L}
\newcommand{\kmem}{K^{\textrm{Mem}}}
\newcommand{\vmem}{V^{\textrm{Mem}}}
\newcommand{\qmem}{Q^{\textrm{Mem}}}
\newcommand{\amem}{A^{\textrm{Mem}}}
\newcommand{\mdim}{d_{\textrm{model}}}
\newcommand{\cat}{\textrm{Concatenate}}
\newcommand{\attention}{\textrm{Attention}}

\newcommand{\Kt}{\tilde{K}}
\newcommand{\Vt}{\tilde{V}}

\def\eqref#1{equation~\ref{#1}}

\def\1{\bm{1}}

\DeclareMathAlphabet{\mathsfit}{\encodingdefault}{\sfdefault}{m}{sl}
\SetMathAlphabet{\mathsfit}{bold}{\encodingdefault}{\sfdefault}{bx}{n}

\newcommand{\R}{\mathbb{R}}

\newcommand{\softmax}{\mathrm{softmax}}



%% file: sections/abstract.tex
\begin{abstract}
The ``state'' of State Space Models (SSMs) represents their memory,  which fades exponentially over an unbounded span. 
By contrast, Attention-based models have ``eidetic'' (i.e., verbatim, or photographic) memory over a finite span (context size). Hybrid architectures combine State Space layers with Attention, but still cannot recall the distant past and can access only the most recent tokens eidetically.
Unlike current methods of combining SSM and Attention layers, we allow the state to be allocated based on relevancy rather than recency. In this way, for every new set of query tokens, our models can ``eidetically'' access tokens from beyond the Attention span of current Hybrid SSMs without requiring extra hardware resources.  
We introduce a method to expand the memory span of the hybrid state by ``reserving'' a fraction of the Attention context for tokens retrieved from arbitrarily distant in the past, thus expanding the eidetic memory span of the overall state. We call this reserved fraction of tokens the ``expansion span,'' and the mechanism to retrieve and aggregate it ``\ourattndef'' (\ourattn).
To adapt Hybrid models to using \ourattn, we propose a novel fine-tuning method that extends LoRA to Hybrid models (\ourlora) and allows efficient adaptation on long spans of tokens.
We show that \ourattnsp enables us to efficiently adapt pre-trained Hybrid models on sequences of tokens up to 8 times longer than the ones used for pre-training.
We show that \ourlorasp with \ourattnsp is cheaper and more performant than alternatives like LongLoRA when applied to Hybrid models on natural language benchmarks with long-range dependencies, such as PG-19, RULER, and other common natural language downstream tasks.
\end{abstract}

%% file: sections/intro.tex
\section{Introduction}
\label{sec:intro}

State Space Models are able to process sequences with an unbounded number of tokens by maintaining a fixed-size state. However, this state is lossy and information about early tokens ``fades'' as more inputs are processed. 
In contrast, Transformer models have a state determined by the number of tokens in their input sequence and are able to access information from all past tokens in their context ``eidetically.'' However, they do so at the cost of extra compute and memory.

Recent Hybrid models \cite{zancato2024b, glorioso2024zamba, de2024griffin} augment SSMs with Attention layers in an effort to counteract SSMs' ``fading'' memory. However, Attention layers only aggregate information by recency (i.e., they process the keys and values of the most recent tokens up to hardware limitations).
Hence, any token that lies beyond the Attention's limited span can only be approximately recalled through the SSMs' state. This limits the effectiveness of Hybrid SSMs when applied to long sequences of tokens, especially when compute and memory resources are limited. 
In fact, while recent pre-trained Hybrid models have demonstrated strong performance on common downstream and recall-intensive language tasks, they are trained following Transformers' pre-training practice: they are either trained on a fixed, relatively short window of tokens when compute is limited---which is often set to 2048 as in LLaMA1 \cite{touvron2023llama}, or 4096 as in LLaMA2 \cite{touvron2023llama2}---or are progressively fine-tuned on longer sequences after the pre-training stage \cite{lieber2024jamba, glorioso2024zamba, dubey2024llama}. 

In this work, we modify Hybrid SSMs' Attention layers to allow their state to be allocated by \textit{relevancy} rather than \textit{recency}, allowing our models to retrieve information that would otherwise fall outside their Attention span.
To do so, we propose \ourattndefsp (\ourattn), a selection mechanism that enables eidetic retrieval of tokens from the past. While \ourattnsp can be applied to both Hybrid SSMs and Transformers, we focus on hybrid architectures where this retrieval capability naturally complements the fading memory of SSMs: while SSMs efficiently process long sequences with their state-based memory, \ourattnsp selectively preserves important information that would otherwise fade in the SSM state.

Our method allows long context fine-tuning of any pre-trained Hybrid model, such as Mamba-2-Hybrid \cite{dao2024mamba2}, Jamba \cite{lieber2024jamba}, and Zamba \cite{glorioso2024zamba2}, on sequences longer than the one used for pre-training without significant compute/memory overhead. 
While Transformer-based fine-tuning strategies for processing long sequences can also be applied to Hybrid models, they typically assume that information in the Attention layers is aggregated by recency and are thus more expensive as sequences get longer.  For example, extending a LLaMA model from a 2k to 8k context size can require up to 32 A100 GPUs \cite{chen2023extending}. While there exist efficient LoRA-based methods \cite{hu2021lora} that lower this cost, we show that Transformer-based methods tend to not work as well for Hybrid models as they do for Transformers. We show that this is mostly due to the fact that combining LoRA (on Attention layers) with SSM layers is not expressive enough to model long-range dependencies beyond the pre-training context size.  

To overcome such limitations, we propose to modify the standard LoRA recipe to fit the Hybrid model structure better. Namely, we employ a fine-tuning strategy similar to LoRA+ \cite{chen2023longlora} on Attention layers, while allowing the SSMs' recurrent parameters to be adapted as well. In particular, building upon previous observations \cite{zancato2024b, yang2024parallelizing_no_conv}, we also adapt the 1D convolutional layers; we empirically demonstrate that this adaptation yields the best results at a reduced cost.

We make the following contributions: (1) We propose \ourattn, a novel selection mechanism that ``reserves'' a fraction of the context of Hybrid SSMs' Attention layers for tokens that are retrieved from the past based on their relevance to the current query. (2) We empirically show that Transformer-based methods for long-context fine-tuning are sub-optimal when applied to Hybrid SSMs. Therefore, we propose \ourlora, an extension of LoRA+ \cite{chen2023longlora} for efficient and effective fine-tuning of Hybrid SSMs. (3) We show that combining \ourattnsp and \ourlorasp on Hybrid models reliably extends their context size up to $8\times$ their pre-training size. Further, we show improved performance over existing adaptation methods on common natural language tasks in the LM Harness Evaluation suite \cite{eval-harness} and long context tasks in RULER \cite{hsieh2024ruler}.

%% file: sections/background.tex
\section{Background and Related Work}

\noindent \textbf{Attention on long contexts.} 
Since the introduction of Self-Attention \cite{vaswani2017attention}, a significant amount of research has been made to reduce its quadratic cost in the sequence length at training time. 
To achieve this, state of the art methods usually approximate the Attention mechanism with sparse/linearized versions.
For example, Reformer \cite{kitaev2020reformer} uses locality-sensitive-hashing to group tokens with similar embeddings, allowing the model to only attend to a subset of tokens rather than the entire sequence. 
Other works have proposed to endow Transformers models with ``compressed'' memory tokens that are updated dynamically and causally over sliding windows on entire sequence chunks. 
For example, Transformer-XL \cite{dai2019transformer} and Infini-Attention \cite{munkhdalai2024leave} segment an input sequence into chunks and process them sequentially while maintaining a complementary set of tokens whose purpose is to summarize the older ones. 
In contrast to these works, our \ourattnsp retrieves relevant tokens ``eidetically,'' i.e., we retrieve tokens rather than maintain a compressed representation of the past. Most similar to our method is Landmark Attention \cite{mohtashami2023landmark}, which inserts landmark tokens into the input at fixed block intervals and trains these tokens to act as summaries of their corresponding blocks via a grouped softmax attention; these summary tokens are then used to index and retrieve relevant input tokens when processing future segments. Our \ourattnsp aims to integrate retrieval natively, without the need for external landmark tokens or complex Softmax procedures. 

\noindent \textbf{State Space Models.} While a great effort has been made to improve the efficiency of Transformer models, a recent line of work has explored efficient alternative ``linear'' architectures. In particular, State Space Models (SSMs) \cite{gu2023mamba, gu2021combining, gu2021efficiently, yang2023gla, sun2023retentive} have emerged as promising competitors to Transformer models due to their efficient scaling and strong empirical performance. 
Numerous variations of SSMs have been proposed, some closely resembling Linear Attention \cite{sun2023retentive} or Linear-Time Invariant dynamical systems \cite{gu2021combining}, while others introduce novel adaptive/gated state updates \cite{yang2023gla, gu2023mamba, dao2024mamba2, orvieto2023resurrecting}. Despite their differences, all follow the same basic working principle that is inspired by classical state space models \cite{kalman1960new}: they process the input sequence by maintaining a \textit{fixed-size} state which acts as a compressed (lossy) representation of all the processed tokens. 
When implemented in hardware, the state must have finite precision and therefore ``fades'' as more samples are processed. 
To overcome this limitation, the most successful SSMs are typically hardware-aware implementations that efficiently utilize modern GPUs/TPUs. These implementations use highly parallelizable and scalable primitives, such as associative scans \cite{gu2023mamba, de2024griffin}, chunking mechanisms \cite{dao2024mamba2, yang2023gla}, and techniques that avoid the materialization of the entire state in slow high bandwidth memory \cite{gu2023mamba}.

\begin{figure*}[ht!]
\begin{center}
\centerline{\includegraphics[width=0.95\textwidth]{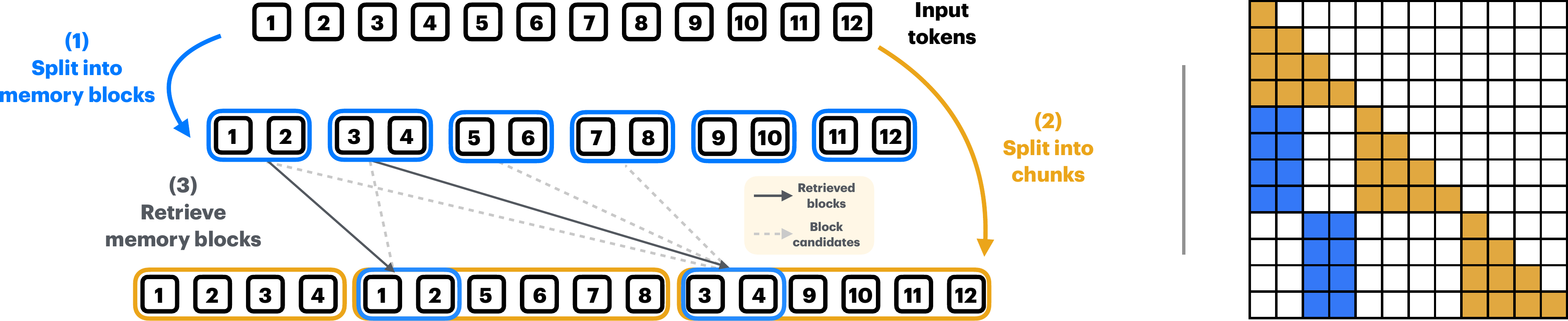}}
\caption{\textbf{\ourattndefsp (\ourattn) overview.} \ourattnsp is a Sparse Attention mechanism used to expand the memory span of Hybrid SSMs. \textit{Left}: \ourattnsp works by reserving a fraction of the Attention context for tokens retrieved arbitrarily far back in the past. We call this reserve the ``expansion span,'' and we populate it with blocks of previous tokens (``memory blocks''). When new tokens arrive, a similarity-based search compares the queries with past memory blocks—represented as summary tokens—to retrieve relevant memory blocks. Then, these retrieved memory blocks are jointly processed with the queries via Attention. While the final Attention mechanism always processes a fixed number of tokens, it can have a longer span since it retrieves tokens from arbitrarily far back in the past. \textit{Right}: Retrieving tokens from the past yields a sparse Attention pattern.
}
\label{fig:teaser}
\end{center}
\vskip -0.3in
\end{figure*}

\noindent \textbf{Hybrid State Space Models.}
While extending the recurrent state in SSMs layers has led to performant models, they typically fall short on tasks that require recalling information from the distant past \cite{waleffe2024empirical, jelassi2024repeat}.  
Hybrid State Space Models have been introduced to solve this limitation and are typically designed to complement the SSMs' ``fading'' state with Attention layers \cite{dao2024mamba2, de2024griffin, lieber2024jamba, glorioso2024zamba}.
While early architectures simply stack SSMs and Attention layers with different blending ratios \cite{waleffe2024empirical, gu2023mamba, dao2024mamba2} or replace full Attention layers with Sliding Window Attention \cite{de2024griffin}, more performant and sophisticated hybrid architectures have been recently proposed \cite{glorioso2024zamba, zancato2024b}. 
In particular, B'MOJO \cite{zancato2024b} complements SSMs' fading state with a form of ``eidetic'' memory whereby SSM layers are mixed with Sliding Window Attention and tokens in the window can attend to a set of tokens from the past that are deemed difficult to predict using an asynchronous causal selection mechanism. 
Similarly, in our work, we mix SSM layers with an Attention layer that operates over chunks of tokens, but we enable these chunks to dynamically retrieve tokens from the past that are most relevant to the chunk's tokens, rather than tokens selected a priori as in B'MOJO.

%% file: sections/method.tex
\section{\ourattndefsp}

Our goal is to enable pre-trained Hybrid SSMs to accurately process sequences longer than the ones seen during pre-training. For a sequence of $\seqlen$ tokens with model dimension $\mdim$, the computational complexity of standard Self-Attention is $O(\mdim \seqlen^2)$. For very large $\seqlen$, computing Self-Attention can be prohibitively expensive. Training models with a large $\seqlen$ is particularly challenging, as training has the additional cost of computing gradients, further limiting memory resources. To address this, we propose \ourattndefsp (\ourattn), a drop-in replacement for standard Self-Attention in Hybrid SSMs. To train \ourattn, we propose \ourlora, a variant of LoRA+ \cite{chen2023longlora} specifically tailored to Hybrid models. A schematic of our \ourattnsp is provided in \cref{fig:teaser}.

\subsection{Attention}
At the heart of modern LLMs is the Attention mechanism, whose role is to construct a contextualized representation of its input. The input to Attention, $x \in \R^{\seqlen \times d}$, is a tensor of $\seqlen$ tokens, each with dimension $d$. Attention is parameterized by  $W_Q, W_K, W_V \in \R^{d\times \mdim}$ and $W_o \in \R^{\mdim \times d}$, which are used to construct linear projections of the input: $Q=xW_Q$, $K=xW_K$, $V=xW_V \in \R^{L \times \mdim}$. After adding positional information to the keys and queries, the output of the Attention layer is\footnote{For simplicity, we consider Single-Head Attention in this exposition.} $\attention(Q, K, V) = \left[\softmax\left(\frac{QK^T}{\sqrt{\mdim}}\right)V\right]W_o \in \R^{L\times d}$. The realization of the attention score matrix, $QK^T \in \R^{L \times L}$, grows quadratically in the sequence length, $\seqlen$. Moreover, the output of Attention is typically followed by a feed-forward layer (FFN), which expands and contracts the dimension of the input. While FFNs are generally regarded as being the primary computational bottlenecks of Attention-based models for short contexts, when the sequence length exceeds the expanded dimension, Attention becomes the primary bottleneck. Since LLMs consist of many layers of Attention, this computation is particularly expensive during training, where activations are cached. To make the Attention mechanism computationally amenable to longer contexts during training, we draw inspiration from RAG and Sparse Attention methods.

\vspace{-0.3cm}
\subsection{Amnesic Attention}\label{sec:hydra_no_mem}
The crux of our  method is in the chunking of the $\seqlen$ tokens in the input sequence $x$, into chunks of size $M$. Namely, we begin by computing projections $Q=xW_Q$, $K=xW_K$, $V=xW_V$ and adding positional information as in standard Attention. Next, each of the $Q, K, V \in \R^{L \times \mdim}$ projections are split into $T = \frac{L}{M}$ chunks of $M$ tokens in each along the sequence dimension, yielding $Q_i, K_i, V_i \in \R^{M \times \mdim}$ for $i=1, \ldots, T$. A naive way to reduce the quadratic cost of Attention is to apply it independently on each of these chunks, $A_i = \attention(Q_i, K_i, V_i)$ and then concatenate and project them to get the final output, $\cat(A_1, A_2, \ldots, A_T)$. However, since the chunks are processed independently, there is no information exchanged between them. Hence, this naive---though efficient---Attention mechanism, which we refer to as ``\ourattn-NoMem'', cannot model time dependencies on contexts larger than the chunk size. 

\vspace{-0.2cm}
\subsection{Eidetic Retrieval Attention}\label{sec:hydra_retrieval}
In this section, we improve upon \ourattn-NoMem by allowing different chunks to exchange information while minimizing compute; we call this Attention mechanism \ourattn.
To this end, we augment the processing of each chunk with a mechanism to retrieve tokens from previous chunks. In particular, we allow chunk $i$ to retrieve tokens from  chunks $1, 2, \ldots, i-1$ and, when the most relevant chunks are selected, append their tokens to its context (its ``expansion span'').
\ourattnsp takes as input a sequence, $x \in \R^{L\times d}$ and computes projections $Q=xW_Q$, $K=xW_K$, $V=xW_V$ followed by the addition of RoPE \cite{su2024roformer} embeddings as in standard Attention. As described in \cref{sec:hydra_no_mem}, $Q, K, V$ are split into $T$ chunks with $M$ tokens in each, yielding tuples $(Q_i, K_i, V_i)$, $i=1, \ldots, T$. Additionally, $Q, K, V$ are split into a second set of $U$ chunks with $S$ tokens in each, yielding tuples $(\qmem_j, \kmem_j, \vmem_j)$ where $\qmem_j, \kmem_j, \vmem_j \in \R^{S\times \mdim}$ for $j=1, \ldots, U$. We refer to these tuples as ``memory blocks''; in \ourattn, the query from each chunk, $Q_i$, attends not only to $K_i$, but also to a set of retrieved memory blocks which populate \ourattn's expansion span. In particular, each $Q_i$ retrieves $\topk$ (top-$\topk$) key/value memory blocks from the past\footnote{By ``the past,'' we mean tokens in the sequence that came before tokens in the chunk.}: $\left(\kmem_{\phi_i(U)_1}, \vmem_{\phi_i(U)_1}\right), \ldots, \left(\kmem_{\phi_i(U)_\topk}, \vmem_{\phi_i(U)_\topk}\right)$ where $\phi_i(U)_j$ denotes the index of the $j$-th memory block selected by the $i$-th chunk. Retrieved blocks are appended to the chunks' keys and values, and \ourattnsp computes the Attention output for the $i$th chunk as in \cref{eq:hydra_att}.
\begin{align}
    \label{eq:mem_cat_k} \Kt & = \cat(\kmem_{\phi_i(U)_1}, \ldots, \kmem_{\phi_i(U)_\topk}, K_i) \\
    \label{eq:mem_cat_v} \Vt &= \cat(\vmem_{\phi_i(U)_1}, \ldots, \vmem_{\phi_i(U)_\topk}, V_i) \\
    \label{eq:hydra_att} A_i^{\textrm{\ourattn}} &= \attention(Q_i, \Kt_i, \Vt_i)
\end{align}
Afterwards, each chunk's Attention outputs are concatenated and projected:
\begin{align}\nonumber
    o^{\textrm{\ourattn}} &= \cat(A_1^{\textrm{\ourattn}}, A_2^{\textrm{\ourattn}}, \dots, A_T^{\textrm{\ourattn}}).
\end{align}
Note that \cref{eq:hydra_att} is a form of Cross-Attention since we are not concatenating memory query tokens to the chunk's query.
This is to preserve causality, as the retrieved tokens cannot attend to the chunk's tokens. 

\noindent \textbf{Memory Retrieval.} Each chunk $x_i$ must judiciously select which memory blocks to retrieve from the past. To do this efficiently, we associate a 1-dimensional tensor, $c_j\in \R^{\mdim}$, to each of the $j=1, \ldots, U$ memory blocks which act as a compressed representation of each memory block. To determine which memory blocks $Q_i$ should attend to, we compute a ``relevancy score'' between $Q_i$ and each $c_j$, which measures how relevant memory block $j$ is to chunk $i$. This relevancy score is implemented with a Cross-Attention score between chunks and compressed memory block representations. Recalling that $Q_i \in \R^{M \times \mdim}$, we compute relevancy score $R_{ij} \in \R$ between chunk $i \in \{1, 2, \ldots, T\}$ and memory block $j \in \{1, 2, \ldots, U\}$ as follows: 
$R_{ij} = \sum_{t=1}^M (Q_ic_j)_t.$
$R_i \in \R^U$ represents the relevancy score between chunk $i$ and all memory blocks. However, since chunk $i$ should only retrieve memory blocks that came before it temporally, we add a mask to $R_i$ and then apply Softmax to obtain the final scores, $\tilde{R}_i$, between chunk $i$ and all memory blocks as follows:
\begin{align}\label{eq:sm_relevancy_score}
 \tilde{R}_i &= \softmax
\left(
\frac{1}{\sqrt{\mdim}}(R_i + \mathcal{M}_i)
\right) \in \R^{U}
\end{align}
where $\mathcal{M}_i$ is a mask of 0 and $-\infty$ constructed to set the scores of future memory blocks (relative to chunk $i$) to $-\infty$.  Once all relevancy scores are computed, chunk $i$ simply retrieves the top $\topk$ memory blocks with the highest $\tilde{R}_{ij}$ scores and concatenates them with the keys and values as in \cref{eq:mem_cat_k,eq:mem_cat_v} before computing Attention as in \cref{eq:hydra_att}.

\noindent \textbf{Compressed Memory Blocks.} Next, we discuss how we construct the compressed memory block representations, $c_j$. Recently, Landmark Attention \cite{mohtashami2023landmark} considered using ``landmark'' tokens to obtain compressed representations of memory blocks. We consider using landmark tokens to construct $c_j$ in \cref{sec:landmark_ablation}. In \ourattn, we consider a simpler approach which we found to work well. For each memory block, $(\qmem_j, \kmem_j, \vmem_j)$, we perform standard non-causal Attention, $\amem_j = \softmax\left(\frac{\qmem_j(\kmem_j)^T}{\sqrt{\mdim}}\right)\vmem_j \in \R^{S \times \mdim}$; we consider non-causal Attention as we are interested in a global representation where all tokens within the memory block can attend to each other. We compute the mean of these weighted memory tokens as the compressed representation:
$c_j = \frac{1}{S}\sum_{t=1}^S (\amem_j)_t \in \R^{\mdim}$,
where $(\amem_j)_t$ denotes the $t$-th row of $\amem_j$. 

\subsection{Training \ourattnsp with LoRA}
In this paper, we fine-tune models pre-trained with standard Attention; however, we fine-tune them with \ourattn, which modifies the pre-trained Attention mechanism by introducing a retrieval mechanism. \ourattnsp repurposes the model's Attention parameters $(W_Q, W_K, W_V, W_o)$ to perform retrieval. In order to efficiently train the model to learn to use \ourattn, we use a variant of LoRA \cite{hu2021lora}.  Recently, \cite{chen2023longlora} introduced LoRA+ designed to fine-tune Transformer models on long contexts. LoRA fine-tunes Attention parameters with low rank adapter matrices. LoRA+ differs from LoRA by also training embedding and normalization layers. 

Recently, \cite{galim2024parameter} found that pure SSMs can be fine-tuned by training the SSM projection layers with LoRA. Since we fine-tune on long contexts, we prioritize efficient training, and consequently apply LoRA only to the Attention layers of our hybrid models. However, it is common for SSM layers to include a 1D convolution layer after their initial projection in order to perform sequence mixing \cite{zancato2024b, glorioso2024zamba, lieber2024jamba, dao2024mamba2, de2024griffin}. The 1D convolution parameters constitute a very small portion of the model parameters ($\sim$0.7\% for Mamba-2-Hybrid 2.7B), but as discussed further in \cref{sec:mamba_ablations}, we found that training these 1D convolution layers in conjunction with LoRA+ improved our models' performance on long-context tasks; we refer to this augmented LoRA+ variation as \ourlora.

%% file: sections/experiments_no_llama.tex
\vspace{-0.5cm}
\section{Experiments}

\subsection{Experimental Setup}
\textbf{Models.}  We enhance the recall capabilities of pre-trained hybrid SSM models by fine-tuning them with different Attention layers on spans of tokens longer than the ones used for pre-training. We consider Mamba-2-Hybrid 2.7B
\cite{waleffe2024empirical} as our representative SSM hybrid model and provide results for Zamba2-Hybrid \cite{glorioso2024zamba2} in \cref{sec:zamba2}. We also explore expanding the span of Transformer models in \cref{sec:se_attn_on_transformers}. These models, obtained from Hugging Face, were pre-trained with a context size of 2048. In our experiments, ``Non-fine-tuned'' refers to the pre-trained model with no fine-tuning. Similar to \cite{chen2023longlora}, we utilize \ourattnsp for efficient fine-tuning and revert to using \fullattnsp during evaluation; this is discussed further in \cref{sec:eval_with_full}.

\noindent \textbf{Datasets.} We fine-tune models using a common language dataset and provide additional results when fine-tuning on PG-19 \cite{raecompressive2019} and a mixture of language and code dataset in \cref{sec:additional_ft_results}.

\noindent \textbf{Baselines.} We compare \ourattnsp to three Attention variants: standard full Attention (``\fullattn'') \cite{vaswani2017attention}, Sliding Window Attention (``\swattn'') \cite{beltagy2020longformer}, and Shifted Sparse Attention (\ssattn) \cite{chen2023longlora}. \fullattnsp serves as the paragon, as the other methods aim to approximate it. All Attention implementations use FlashAttention-2 \cite{dao2023flashattention}.

\noindent \textbf{Training Procedure.} We adopt the training recipe used in \cite{chen2023longlora} to fine-tune our models, with the exception of using \ourlorasp instead of LoRA+. See \cref{sec:training_details} for more details.

\noindent \textbf{Evaluation Metrics.} We assess the performance of our models across a range of common benchmarks. To assess the predictive capabilities of our fine-tuned models on long sequence lengths, we measure perplexity (PPL) on the PG-19 validation set; however, as discussed further in \cref{sec:additional_lora_ablations}, we do not find PPL to be a faithful measure of a model's long context abilities. To assess performance on more real-world language tasks in the short and long-context settings, we evaluate our models on various LM Evaluation Harness tasks \cite{eval-harness}. To measure our models' performance on long-context tasks, we evaluate on the RULER \cite{hsieh2024ruler} benchmark. We consider eleven RULER tasks, encompassing needle-in-a-haystack, variable-hopping, and aggregation tasks; we aggregate these metrics into five groups as explained in \cref{sec:ruler_aggregation}. We provide additional long-context benchmarks in \cref{sec:appendix_lctx_tasks}.

\vspace{-0.2cm}
\subsection{Results}
All of our Mamba-2-Hybrid models are fine-tuned with a context size of 8192. For \ourattn, we use a block size of 32, and a top-$\topk$ of 8. For \swattn, we use a window size of 4096. \ssattnsp uses the parameters from \cite{chen2023longlora}.
When fine-tuning Mamba-2-Hybrid using \ourattn, we found that applying the same chunk sizes at each layer leads to suboptimal downstream performance. Therefore, we segment each sample into chunks of variable sizes (picked randomly from $\{2048, 4096\}$).
We found this prevents the model from overfitting to a fixed chunk size, which is a hyperparameter chosen a priori, independent of the actual data content.
An ablation on \ourattnsp with different chunk sizes is provided in \cref{sec:mamba_ablations}.

\noindent \textbf{Perplexity.} We provide PG-19 PPL results in \cref{table:mamba_ppl_and_lm_harness_sp}. All fine-tuned models outperform the non-fine-tuned model. \ourattnsp yields the closest performance to the paragon model fine-tuned with \fullattn, outperforming \ssattnsp and \swattnsp across all context sizes at or above the fine-tuning size.

\noindent \textbf{LM Harness.} Next, we evaluate Mamba-2-Hybrid models across short and long-context tasks in the LM Evaluation Harness suite. Our results are provided in \cref{table:mamba_ppl_and_lm_harness_sp}, where we observe that all models perform similarly on short-context tasks---including the non-fine-tuned model---suggesting there is no performance regression when fine-tuning on larger contexts. Furthermore, we observe that fine-tuning with \ourattnsp gives the closest performance to fine-tuning with \fullattn.

\begin{table*}[t]
\centering
\begin{adjustbox}{max width=\textwidth}
\begin{tabular}{c|cccc|ccccccc|cccc}
\specialrule{2.5pt}{1pt}{1pt}
\multirow{2}{*}{\textbf{Attention}} & \multicolumn{4}{c}{\textbf{Eval Context Size (PG-19 PPL $\downarrow$)}} & \multicolumn{7}{c}{\textbf{Short Context Tasks ($\uparrow$)}} & \multicolumn{4}{c}{\textbf{Long Context Tasks ($\uparrow$)}} \\
& 2048 & \cellcolor[gray]{.85} 8192 & 16384 & 32768 & ARC-E & ARC-C & Hella. & LAMB. & PIQA & WG & Avg. & SWDE & SQA & SNQA & Avg. \\
\specialrule{1pt}{1pt}{1pt}
 Non-fine-tuned & 10.72 & 14.99 & 19.35 & 26.37 & 69.91 & 37.97 & 67.62 & 69.84 & 76.06 & 65.04 & \cellcolor[gray]{.9} 64.41 & 85.60 & 15.18 & 3.65 & \cellcolor[gray]{.9} 34.81 \\
\specialrule{1pt}{1pt}{1pt}
\fullattn & 10.99 & 10.28 & 10.39 & 11.14 & 69.53 & 38.48 & 67.30 & 68.93 & 75.08 & 64.40 & \cellcolor[gray]{.9} 63.95 & 85.24 & 26.99 & 19.75 & \cellcolor[gray]{.9} 43.99 \\
\specialrule{1pt}{1pt}{1pt}
\swattn & 10.98 & 10.80 & 11.82 & 13.45 & 69.82 & 38.23 & 67.35 & 69.18 & 75.30 & 63.85 & \cellcolor[gray]{.9} 63.95 & 84.61 & 24.85 & 15.41 & \cellcolor[gray]{.9} 41.63 \\
\hline
\ssattn & 10.87 & 12.89 & 14.67 & 16.37 & 70.12 & 38.05 & 67.39 & 69.84 & 75.95 & 64.56 & \cellcolor[gray]{.9} 64.32 & 86.41 & 17.44 & 8.53 & \cellcolor[gray]{.9} 37.46 \\
\specialrule{1.5pt}{1pt}{1pt}
\ourattn & 10.99 & 10.45 & 11.14 & 12.64 & 70.20 & 38.65 & 67.15 & 69.11 & 75.57 & 63.93 & \cellcolor[gray]{.9} 64.10 & 85.96 & 26.70 & 18.04 & \cellcolor[gray]{.9} 43.57 \\
\hline
\end{tabular}
\end{adjustbox}
\caption{\textbf{Fine-tuning Mamba-2-Hybrid with \ourattnsp outperforms fine-tuning with \ssattnsp and \swattnsp on long-context natural language tasks.} 
We evaluate PG-19 validation perplexity (PPL) and observe that fine-tuning with \ourattnsp preserves performance at longer contexts better than \ssattnsp and \swattn. On long context tasks from the LM Harness suite, \ourattnsp outperforms \ssattnsp and \swattn. See \cref{sec:task_abbrev} for task abbreviation definitions.}
\label{table:mamba_ppl_and_lm_harness_sp}
\vspace{-0.5cm}
\end{table*}

\begin{figure*}[h]
    \centering
    \subfigure[Mamba-2-Hybrid Attention Variants]{\includegraphics[width=0.4\textwidth]{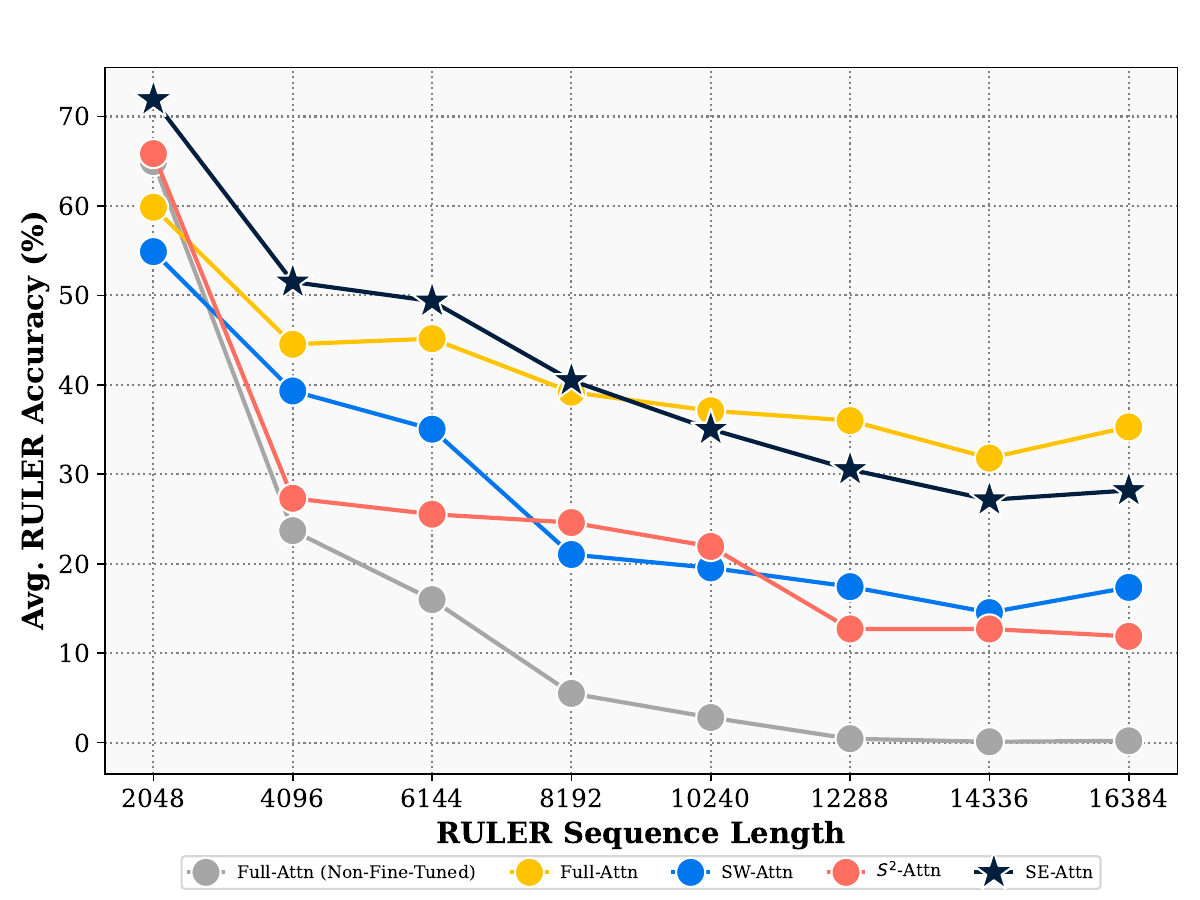}}
    \hfill
    \subfigure[Mamba-2-Hybrid LoRA Variants]{\includegraphics[width=0.4\textwidth]{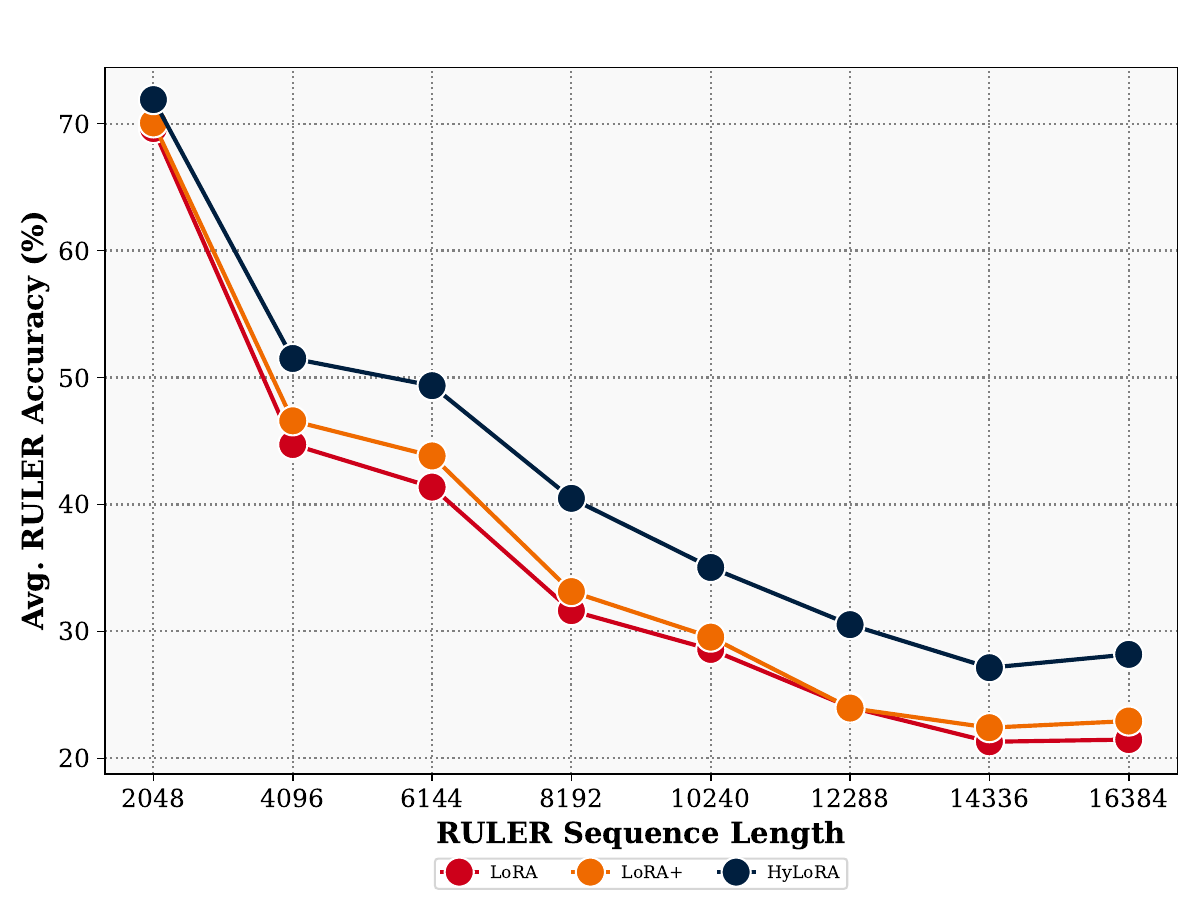}}
    \caption{\textbf{Fine-tuning with \ourattnsp outperforms \swattnsp and \ssattnsp on the RULER benchmark. \ourlorasp outperforms LoRA and LoRA+ on Hybrid models.} (a): We fine-tune Mamba-2-Hybrid with a context size of 8192 and evaluate on eleven RULER tasks, as explained in \cref{sec:ruler_aggregation}. Fine-tuning with \ourattnsp consistently outperforms \swattnsp and \ssattnsp even when evaluating on context sizes beyond the fine-tuning size. (b): We fine-tune Mamba-2-Hybrid with \ourattnsp using LoRA, LoRA+, and \ourlora. LoRA and LoRA+ perform sub-optimally. Our \ourlorasp  additionally trains the 1D convolution layers and yields strong performance.}
    \label{fig:mamba_ruler_avg_hylora}
\end{figure*}

\noindent \textbf{RULER.} Next, we assess the performance of our models on long-context tasks using the RULER benchmark. \cref{fig:mamba_ruler_avg_hylora} shows the average accuracy across eleven RULER tasks. We observe that fine-tuning with \ourattnsp performs similar to fine-tuning with \fullattn, and outperforms \ssattnsp and \swattn. We provide more detailed RULER results in \cref{fig:mamba_ruler_sp}, where we see a substantial improvement on the variable tracking (VT) task, which may be attributed to \ourattn's retrieval during fine-tuning.

\begin{figure*}[h]
    \centering
    \subfigure[]{\includegraphics[width=0.23\textwidth]{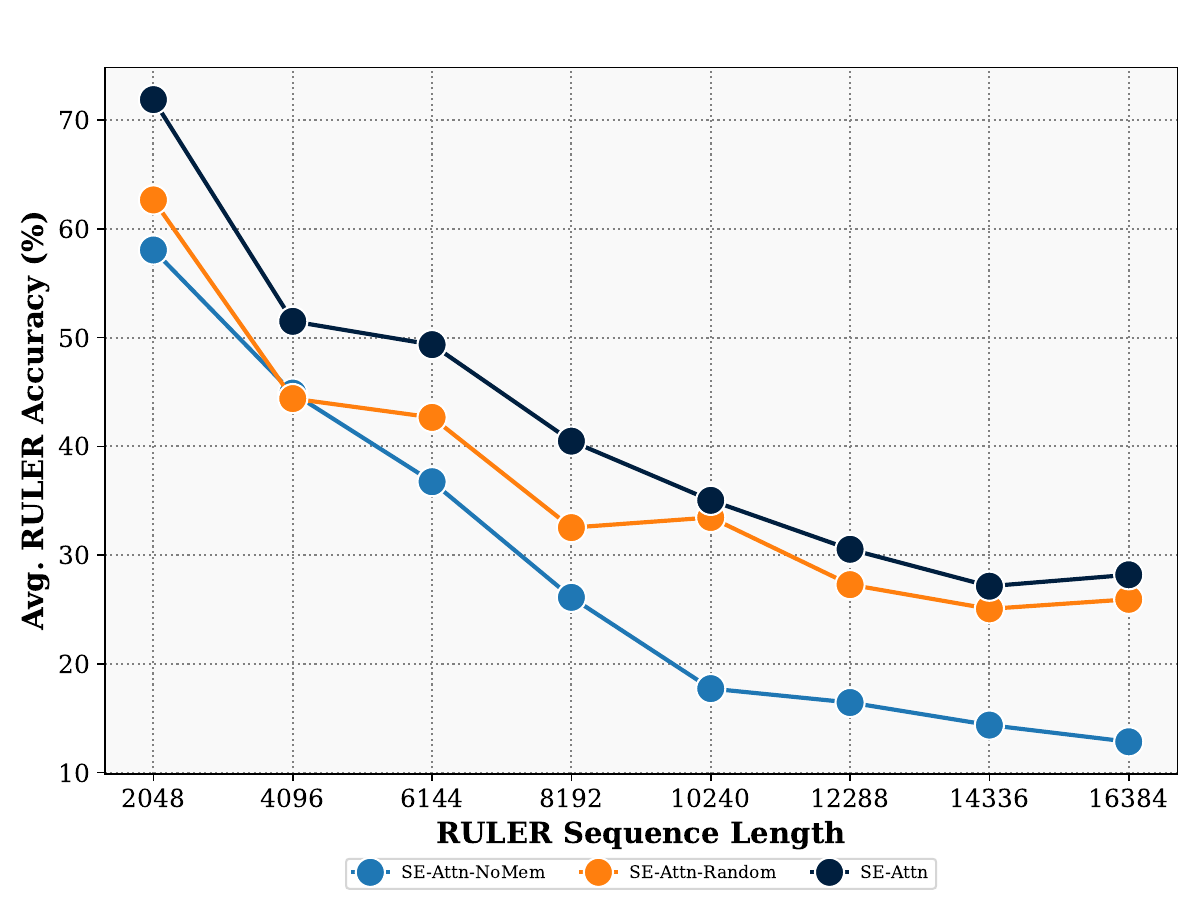}}
    \subfigure[]{\includegraphics[width=0.23\textwidth]{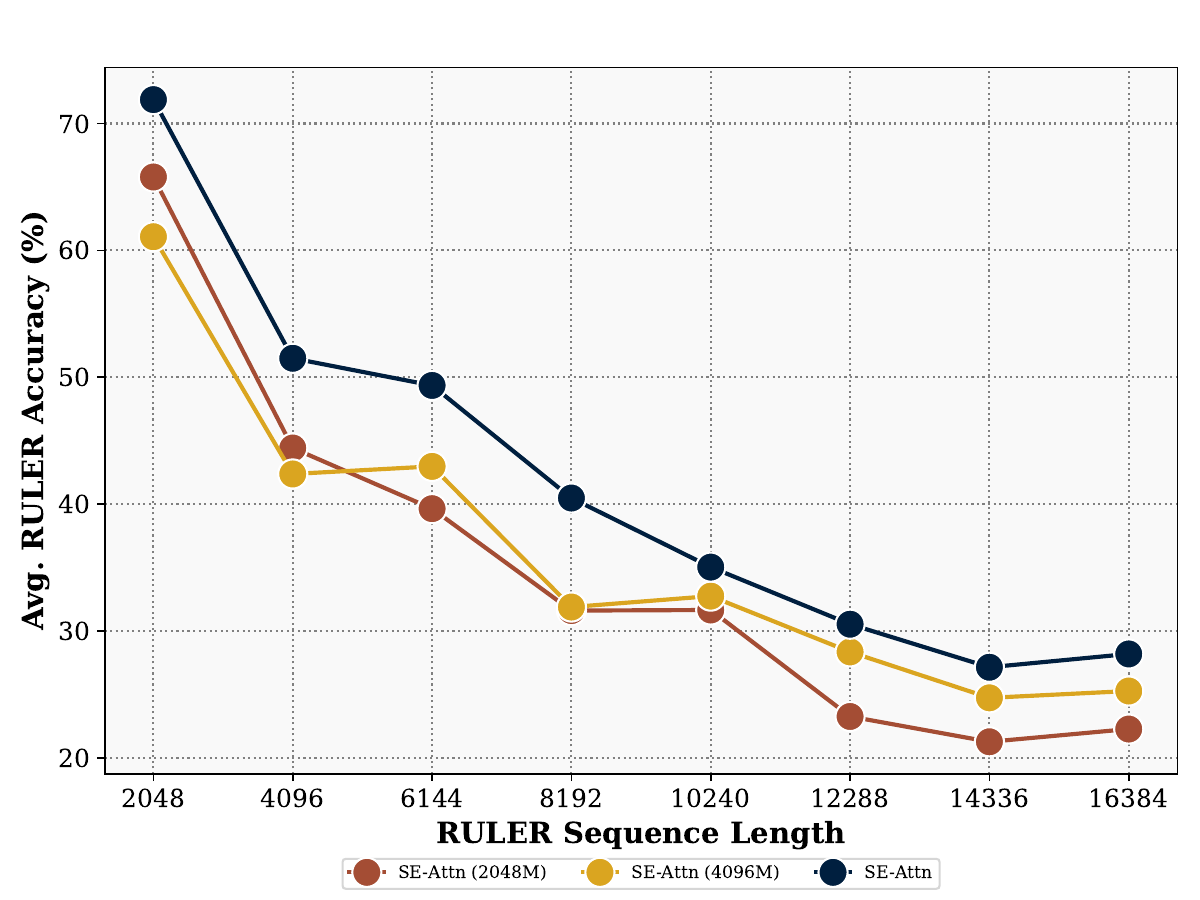}}
    \subfigure[]{\includegraphics[width=0.23\textwidth]{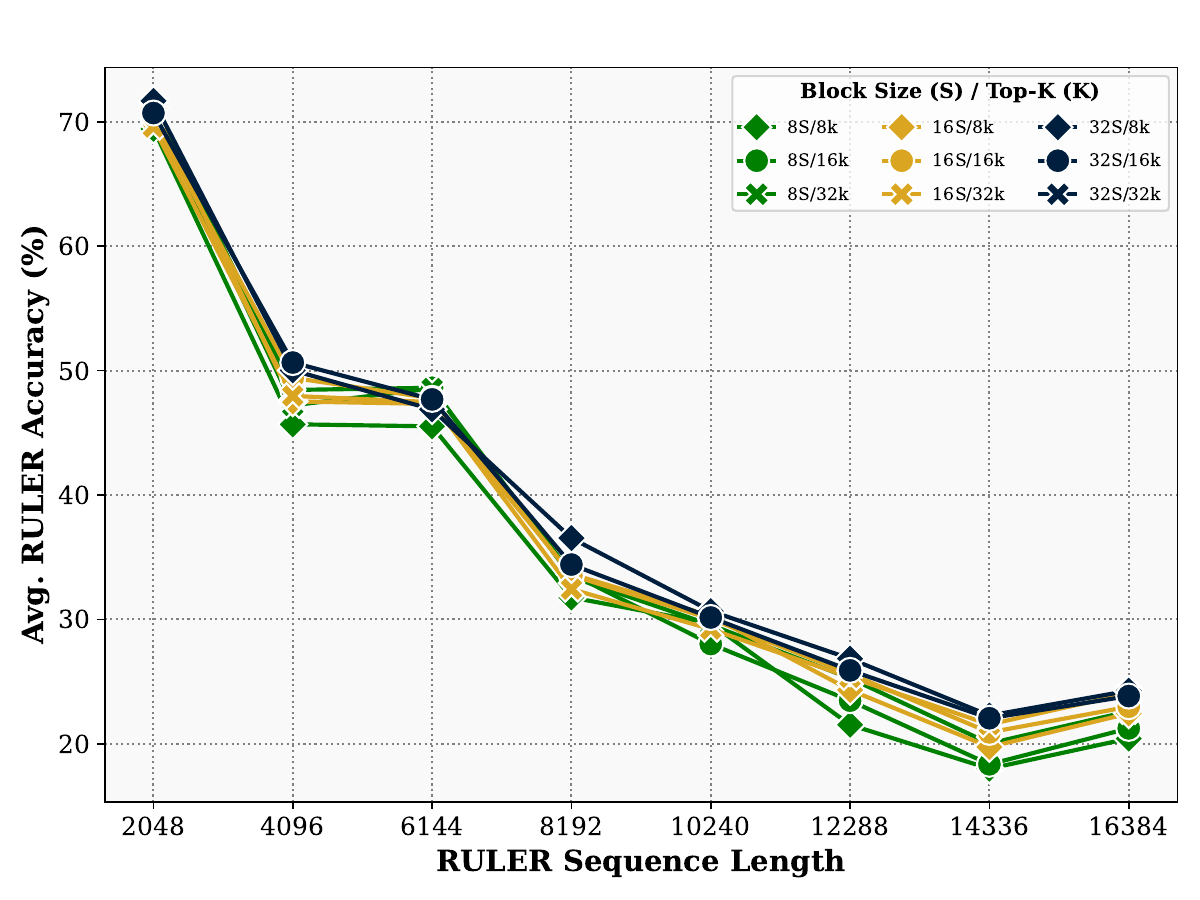}}
    \subfigure[]{\includegraphics[width=0.23\textwidth]{figures/mamba/ablations/sk_comparison.pdf}}
    \caption{\textbf{\ourattnsp ablations on Mamba-2-Hybrid.} (a): Attention-based memory retrieval (\ourattn) improves upon no retrieval and random retrieval. (b): Using \ourattnsp with a chunk size chosen randomly from $\{2048, 4096\}$ acts as a regularizer and outperforms \ourattnsp with fixed chunk sizes of 2048 and 4096. (c): \ourattnsp with larger memory blocks (i.e., more tokens per block) with a smaller top-$\topk$ tends to do better than smaller blocks with a larger top-$\topk$. (d): An expansion span consisting of 256 total tokens (8 memory blocks with 32 tokens in each) gives the strongest performance.
    }
    \label{fig:mamba_hydra_ablations}
\end{figure*}

\vspace{-0.2cm}
\subsection{Ablations}\label{sec:mamba_ablations}

In this section, we ablate over some of the design choices of \ourattnsp on Mamba-2-Hybrid.
For this analysis, we consider the RULER benchmark, as it is a strong indicator of performance on long-context tasks. We use the average of the eleven RULER tasks defined in \cref{sec:ruler_aggregation}.

\noindent \textbf{Does retrieval during training help?} In \cref{sec:hydra_no_mem}, we introduced \ourattn-NoMem, a variant of \ourattnsp where we do not do any retrieval and process chunks independently. Naturally, we do not expect this to do well due to the lack of shared information across chunks. We confirm this in \cref{fig:mamba_hydra_ablations}(a), where we observe that \ourattn-NoMem achieves a much lower performance than \ourattnsp with retrieval. Furthermore, we also fine-tune with \ourattn-Random, a variant of \ourattnsp that populates its expansion span by retrieving random memory blocks for each chunk. This improves upon \ourattn-NoMem, indicating that retrieving some information from the past improves performance; however, it does not do as well as \ourattn, which retrieves blocks based on relevancy.

\noindent \textbf{Chunk size.} We found that using a random chunk size during each forward pass improved upon having a fixed chunk size.  In \cref{fig:mamba_hydra_ablations}(b), we compare the performance of \ourattn, which chooses a random chunk size in $\{2048, 4096\}$ during each layer's forward call, to \ourattnsp with fixed chunk sizes of 2048 and 4096. 
Random chunk sizes outperform fixed ones, suggesting a regularizing effect that makes the model more robust to different context lengths.

\begin{figure*}[h]
    \centering
    \subfigure[]{\includegraphics[width=0.32\textwidth]{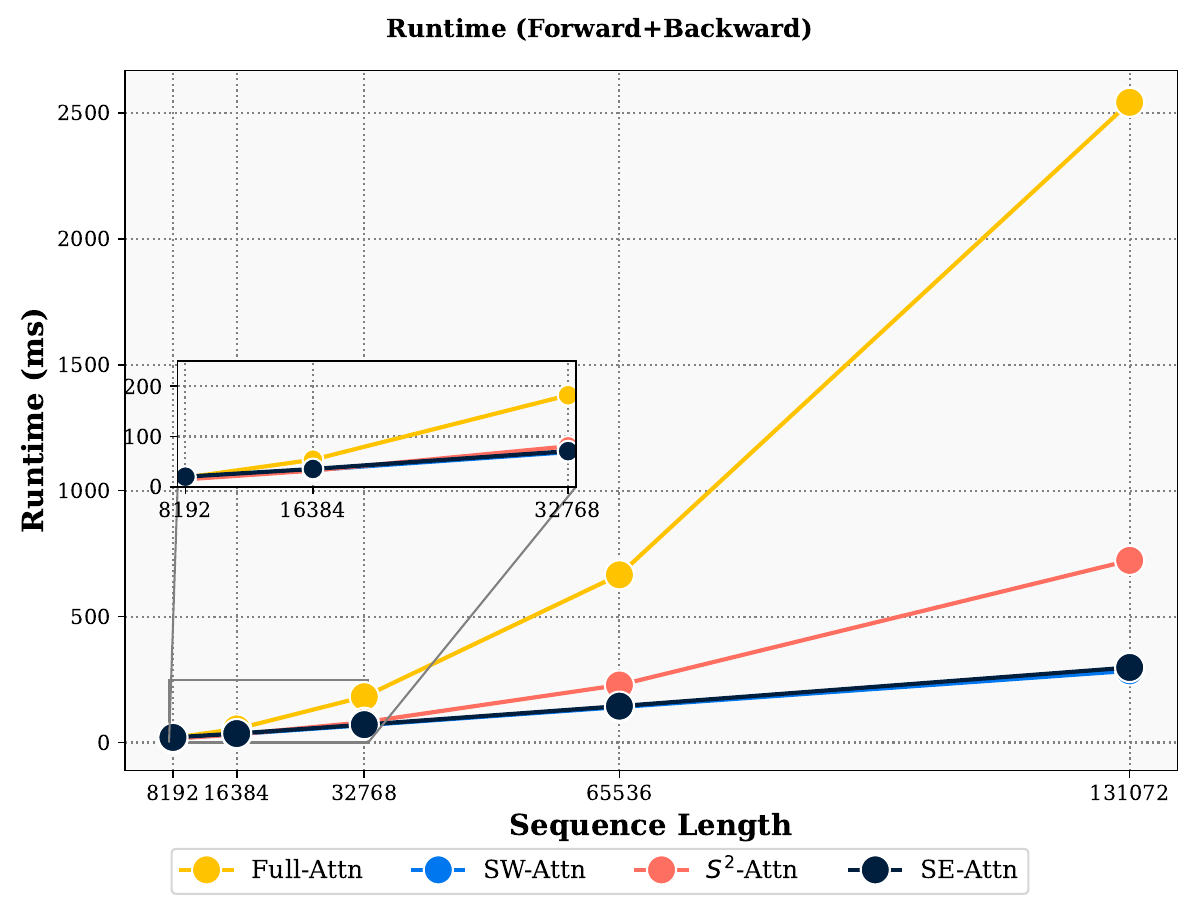}}
    \subfigure[]{\includegraphics[width=0.32\textwidth]{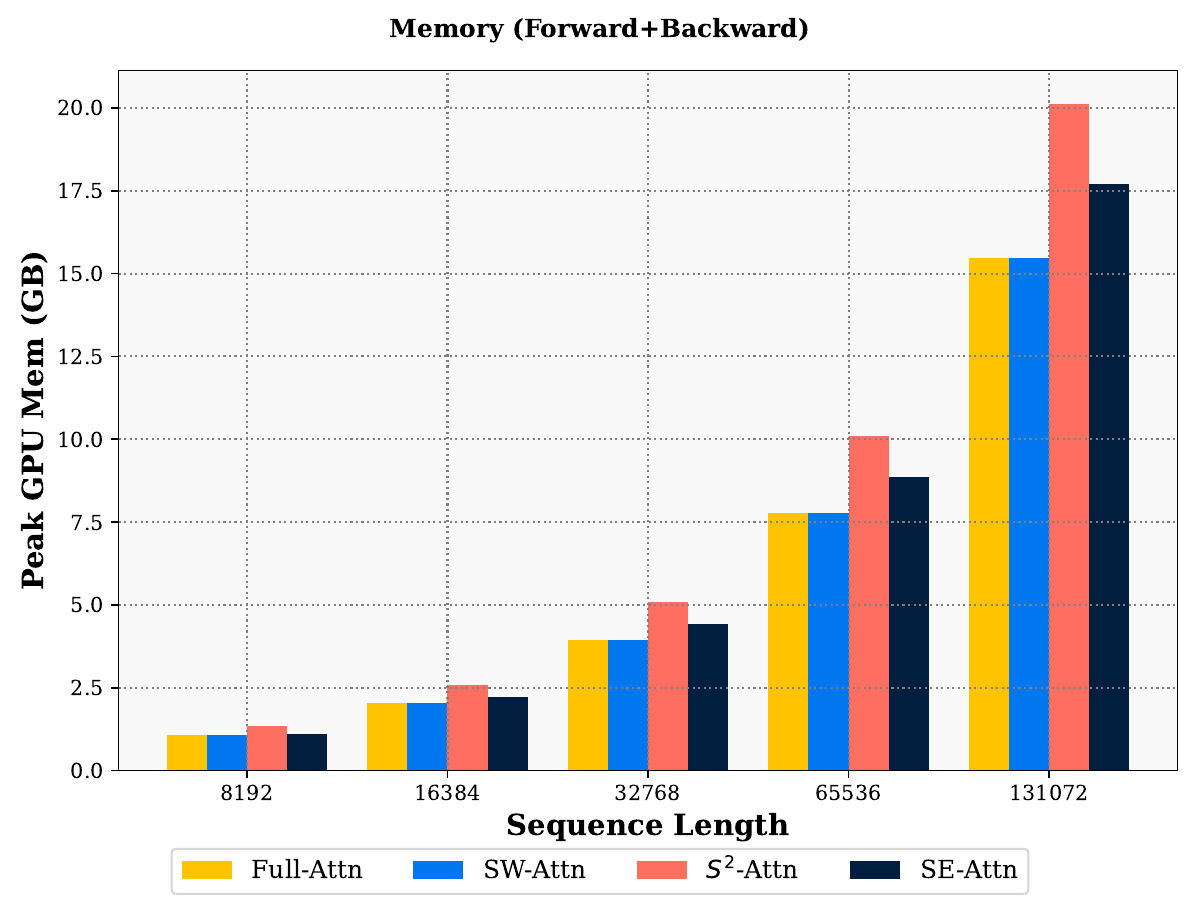}}
    \subfigure[]{\includegraphics[width=0.32\textwidth]{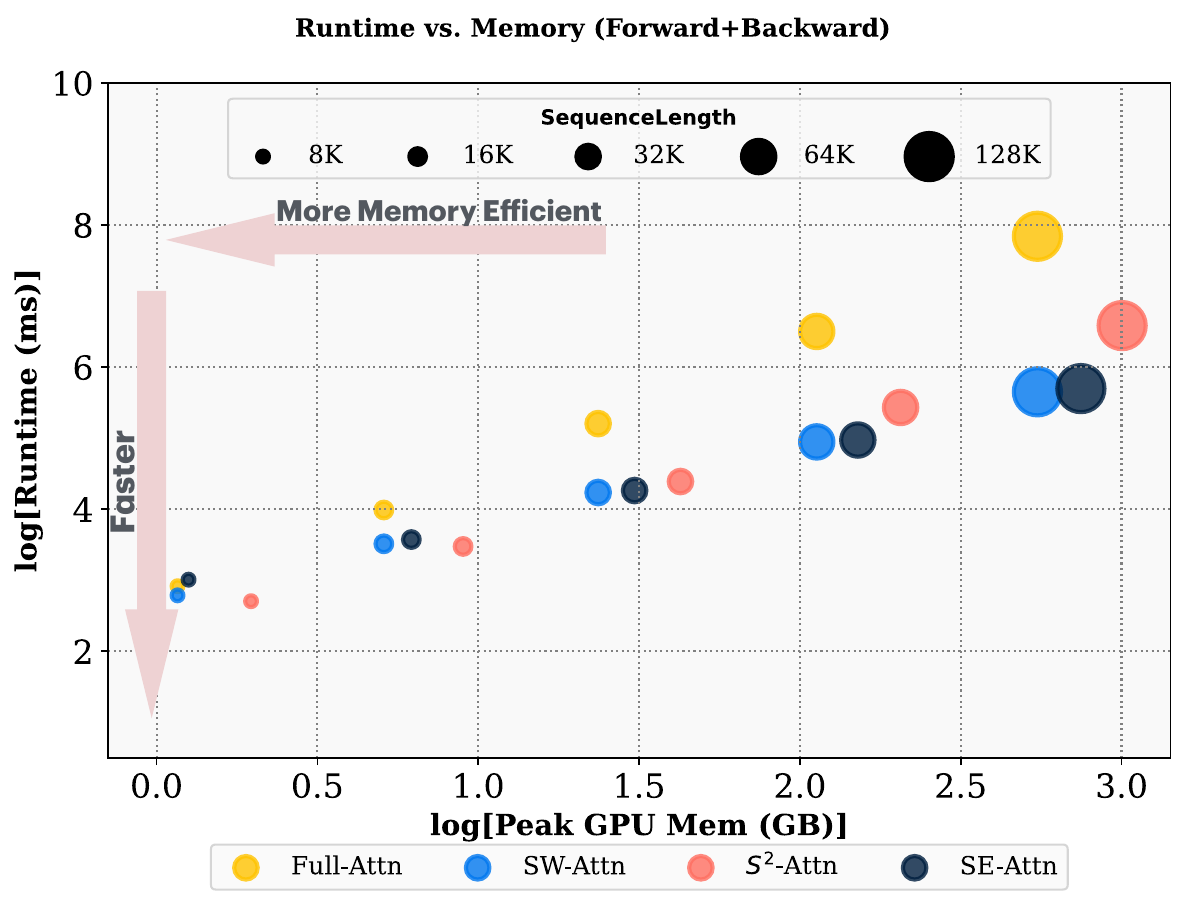}}
    \caption{\textbf{\ourattnsp offers a greater runtime-memory trade-off than other Attention variants.} We profile various attention layers during a single training step. \swattnsp uses a window size of 4096 and \ourattnsp alternates between a chunk size of 2048 and 4096. (a): Runtime of a single training step. (b): Peak GPU memory used during the training step. (c): Runtime vs. memory used. Points on the lower left of the plot exhibit a stronger runtime-memory trade-off.}
    \label{fig:mamba_fwd_bwd_anal}
    \vspace{-0.2cm}
\end{figure*}

\noindent \textbf{Block size and top-$\topk$.} For a fixed expansion span size, we might expect that retrieving memory blocks at a finer granularity should perform better than retrieving a smaller number of large blocks; this is because retrieving a larger number of small blocks is as flexible as retrieving a small number of large blocks. However, as illustrated in \cref{fig:mamba_hydra_ablations}(c), we found that fine-tuning \ourattnsp with larger block sizes and a smaller top-\topk gave the best performance. 
This is possibly due to the increased complexity of the retrieval task, which gets more difficult as the number of possible blocks to retrieve increases. 
As we are fine-tuning with LoRA, we hypothesize that the model's capacity may not be sufficient to learn to retrieve effectively. As illustrated in \cref{fig:mamba_hydra_ablations}(d), we found that an expansion span populated with 256 retrieved tokens (32 memory blocks with 8 tokens in each) works best.

\noindent \textbf{\ourlorasp for Hybrid Models.} We fine-tune our hybrid models using \ourlora, which builds upon LoRA+ \cite{chen2023longlora} by also training 1D convolution layers. In \cref{fig:mamba_ruler_avg_hylora}(b) we show that, when evaluated on RULER tasks, fine-tuning \ourattnsp models with \ourlorasp gives the strongest performance. In \cref{sec:additional_lora_ablations}, we study the effect of the LoRA rank and apply HyLoRA to \fullattn.

\vspace{-0.4cm}
\subsection{Empirical Runtime Analysis}\label{sec:empirical_runtime}

While hardware-efficient Attention implementations, such as FlashAttention \cite{dao2023flashattention}, enable linear memory scaling, computational complexity remains quadratic in sequence length. In \cref{fig:mamba_fwd_bwd_anal}(a), we profile different Attention layers on various context sizes for one training step and  see that \ourattnsp is much faster than \ssattnsp and \fullattn. Moreover, the runtime of \ourattnsp is similar to that of \swattn, despite \ourattn's much longer Attention span.
We also note that the runtime increase of \fullattn/FlashAttention compared to \ourattnsp is 4-5$\times$ for long sequences, while the memory peak increase is 17\%, as shown in \cref{fig:mamba_fwd_bwd_anal}(b). \ourattnsp has a minor memory overhead because it requires storing partial block summary statistics in order to retrieve tokens from the past. However, this overhead is outweighed by its faster runtime. As depicted in \cref{fig:mamba_fwd_bwd_anal}(c), our method has a strong runtime-memory trade-off---similar to that of \swattn, but with the added benefit of improved performance on language benchmarks. See \cref{sec:theoretical_runtime} for a theoretical runtime analysis.

%% file: sections/conclusion.tex
\vspace{-0.4cm}
\section{Conclusion}

We close with the limitations of our method. Although we have conducted experiments at a relatively large model scale (2.7B) and long contexts (up to 32k), testing our method for efficiently adapting larger models on longer contexts is paramount. Furthermore, while we utilize datasets that require modeling of long-range dependencies, we found that perplexity-based tasks do not faithfully measure models' capabilities to handle long contexts, and instead tasks like RULER provide better signals of long-context capabilities. However, RULER is mostly a synthetic dataset and does not cover more nuanced tasks that require reasoning over long documents.  Validating our method on more complex long-range benchmarks is a promising area for future work.

%% file: sections/appendix_llama_overflow.tex
\clearpage
\setcounter{page}{1}

\section{Evaluating with Full Attention}\label{sec:eval_with_full}

Our models are efficiently fine-tuned using \ourattnsp and use \fullattnsp during evaluation, as in \cite{chen2023longlora}. Due to KV-caching, the complexity of \fullattnsp scales linearly during evaluation, so an efficient Attention layer in this setting is not as crucial as in training. Nevertheless, for the sake of completeness, we begin by evaluating our models with the same Attention layer used during fine-tuning (i.e., we fine-tune and deploy our Hybrid model with \ourattn). In \cref{table:ft_hydra_eval_hydra}, we experiment with models that use \fullattn, \swattn, \ssattn,  and \ourattn; then, we evaluate perplexity on the PG-19 dataset. All models were fine-tuned with a context size of 8192. We observe that \swattnsp is best at preserving the perplexity on context sizes up to $32\times$ larger than the one used for pre-training. \ourattnsp is also able to maintain a lower perplexity across longer context sizes, while \ssattnsp deteriorates much more quickly. As discussed further in the next section, perplexity may not be the most faithful metric for assessing models' long-context capabilities.

\begin{table}[t]
\centering
\resizebox{0.5\columnwidth}{!}{%
\begin{tabular}{c|cccc}
\specialrule{2.5pt}{1pt}{1pt}
\multirow{2}{*}{\textbf{Attention}} & \multicolumn{4}{c}{\textbf{Evaluation Context Size (PPL $\downarrow$)}} \\
 & \cellcolor[gray]{.85} 8192 & 16384 & 32768 & 65536 \\
\specialrule{1.5pt}{1pt}{1pt}
Non-fine-tuned & 14.99 & 19.35 & 26.37 & 34.51 \\
\specialrule{1pt}{1pt}{1pt}
\fullattn & 10.28 & 10.39 & 11.14 & 12.38 \\
\hline
\swattn & 10.33 & 10.22 & 10.16 & 10.16 \\
\hline
\ssattn & 10.73 & 10.76 & 11.85 & 13.72 \\
\specialrule{1.5pt}{1pt}{1pt}
\ourattn & 10.47 & 10.70 & 10.91 & 10.96 \\
\hline
\end{tabular}
}
\caption{\textbf{Mamba-2-Hybrid fine-tuned with \ourattnsp or \swattnsp preserves perplexity up to $32\times$ the pre-training context size.} We fine-tune a Mamba-2-Hyrbid model (pre-trained on a context size of 2048) on a context size of 8192 using various Attention variants. We test on longer context sizes using the same Attention used during adaptation.}
\label{table:ft_hydra_eval_hydra}
\end{table}

\section{LoRA for Hybrid Models and the Pitfalls of Perplexity}\label{sec:additional_lora_ablations}
In this section, we conduct a more thorough ablation study on the effect of training Mamba-2-Hybrid with different variants of LoRA, and provide further evidence that perplexity is not a reliable indictor of long context performance.

\begin{figure}[h!]
\vskip 0.2in
\begin{center}
\centerline{\includegraphics[width=0.6\columnwidth]{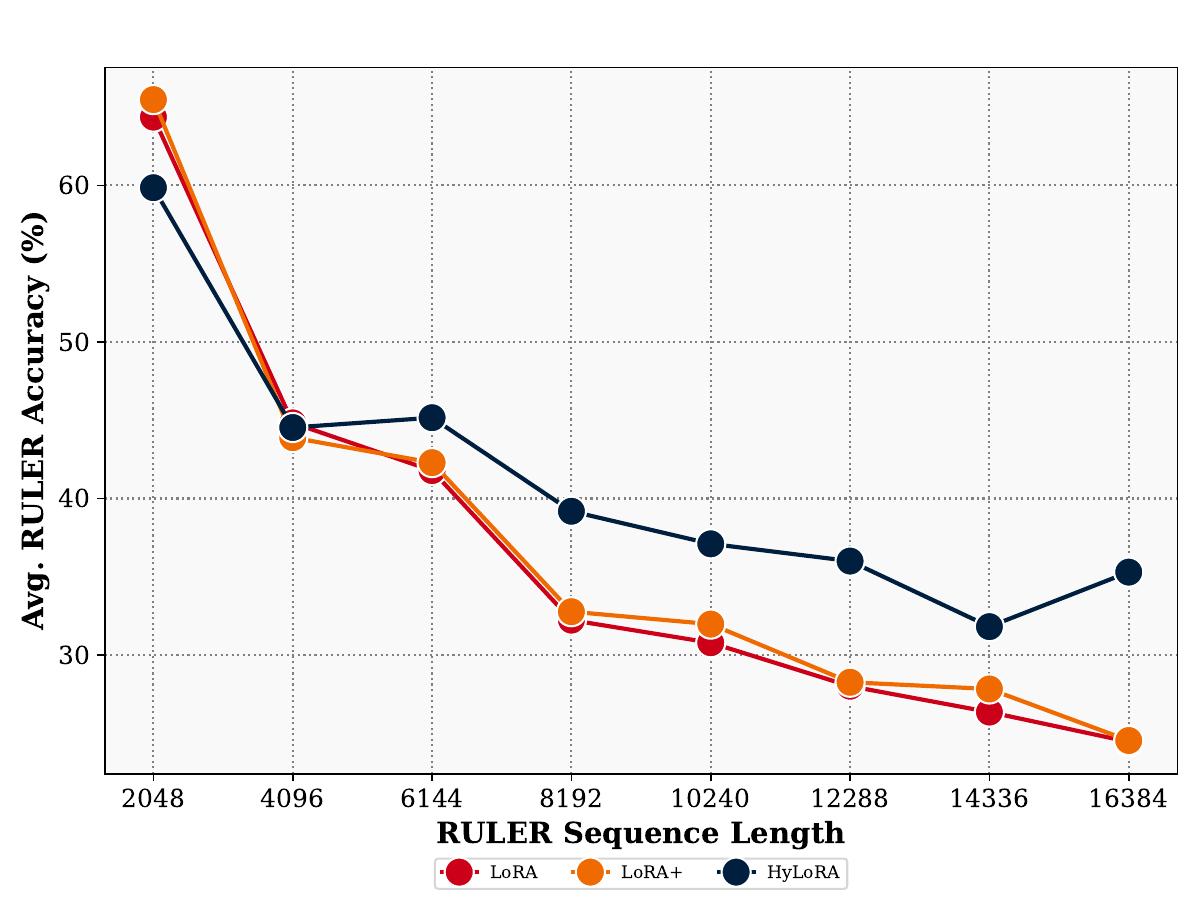}}
\caption{\textbf{\ourlorasp outperforms LoRA and LoRA+ on Hybrid models.} We fine-tune Mamba-2-Hybrid with \fullattnsp using LoRA, LoRA+, and \ourlora. We find that LoRA and LoRA+ perform sub-optimally compared to \ourlorasp which also trains 1D convolution layers.}
\label{fig:full_lora_comp}
\end{center}
\vskip -0.2in
\end{figure}

\noindent \textbf{Fine-tuning with \fullattn.} We first expand upon  \cref{fig:mamba_ruler_avg_hylora}(b), where we showed that fine-tuning Mamba-2-Hybrid with \ourattnsp using \ourlorasp produced a stronger model than training with LoRA and LoRA+. In \cref{fig:full_lora_comp}, we provide a similar plot, but using \fullattnsp during fine-tuning rather than \ourattn. We observe a similar trend, where \ourlorasp produces a stronger model than LoRA and LoRA+. In the remaining analyses in this section, we provide results for fine-tuning with \ourattn, as we observed similar trends when fine-tuning with \fullattn.

\begin{figure}[h!]
\vskip 0.2in
\begin{center}
\centerline{\includegraphics[width=0.6\columnwidth]{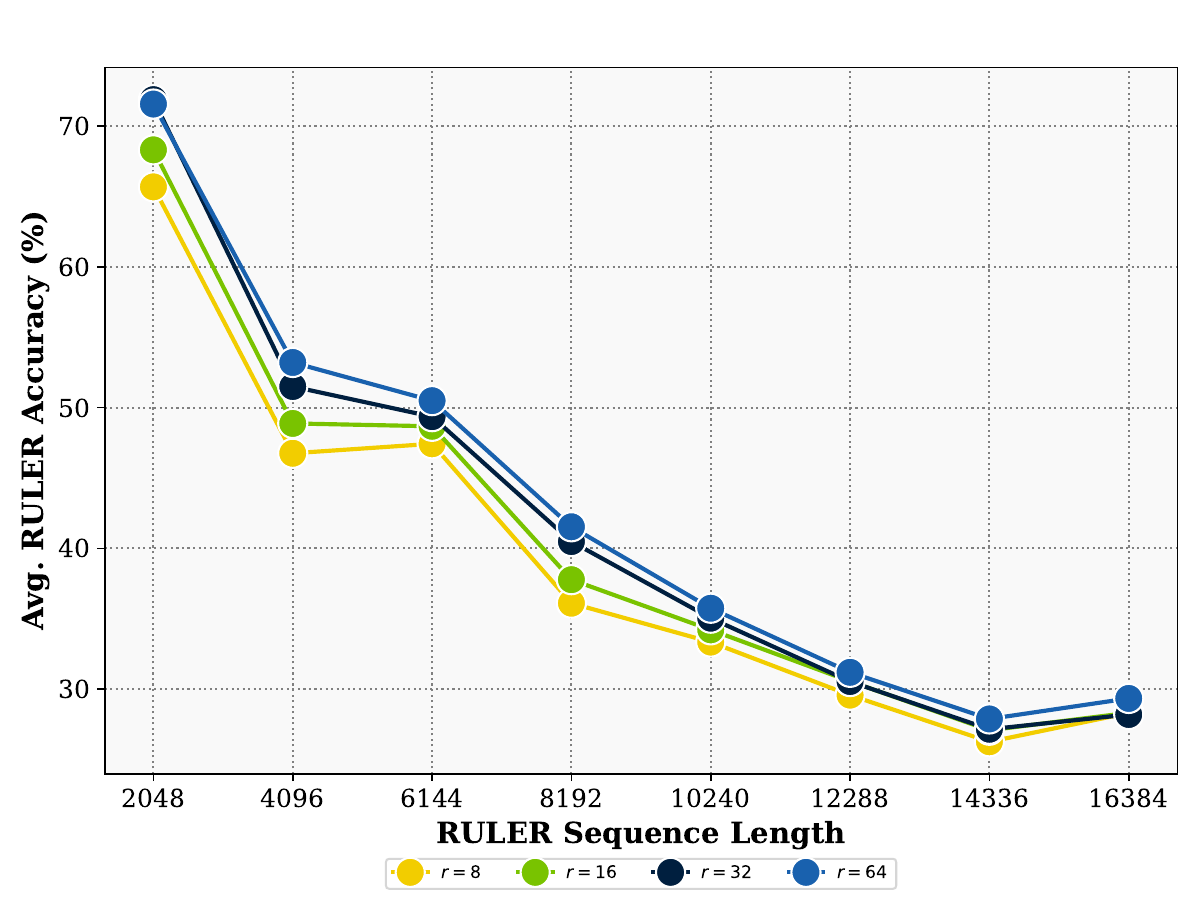}}
\caption{\textbf{Fine-tuning with a larger LoRA rank using \ourlorasp improves performance on the RULER benchmark.} We fine-tune Mamba-2-Hybrid using \ourlorasp with different LoRA ranks (we maintain a LoRA rank to alpha ratio of 2). We observe that fine-tuning with a larger rank produces stronger downstream results on RULER, with some saturation when using a rank of 64.}
\label{fig:lora_rank_comp}
\end{center}
\vskip -0.2in
\end{figure}

\noindent \textbf{Effect of LoRA rank.} We next explore how the rank used for LoRA fine-tuning affects downstream performance on the RULER task. Given its stronger performance, we focus on our \ourlora. In our previous Mamba-2-Hybrid experiments,  we used a LoRA rank $(r)$ of 32 and an $\alpha$ of 64 (for our Transformer experiments, we used $r=8$ and $\alpha=16$ as in \cite{chen2023longlora}). In \cref{fig:lora_rank_comp}, we plot the average RULER results for Mamba-2-Hybrid models fine-tuned with \ourlorasp using different ranks (we scale $\alpha$ to maintain the ratio $\alpha/r = 2$). Here, we observe that training with a larger rank improves downstream performance, and we start to see some saturation around $r=64$. 

In \cref{table:mamba_lora_ablations}, we provide PG-19 perplexity results for Mamba-2-Hybrid trained with different LoRA variants. All models are fine-tuned with a context size of 8192 and evaluated with multiple context sizes. For a given context size, we do not see a substantial difference in the perplexity. This is similar to the observation in \cite{chen2023longlora}. However, as discussed above, the rank can have a significant effect on downstream long-context tasks that require strong recall capabilities, as in RULER. Hence, while perplexity results may be promising, they are not necessarily indicative of performance on more complex long-context tasks. 

Based on the analyses in this section, we conclude the following:

\begin{itemize}
    \item Fine-tuning the 1D convolution layers, as we do in our \ourlora, significantly improves performance on downstream long-context tasks that require retrieval, such as RULER.
    \item Fine-tuning with larger LoRA ranks improves performance up to a certain point---64 for our experiments.
    \item Perplexity may not be the most faithful metric for assessing performance on long-context downstream tasks. Instead, researchers should consider evaluating on more complex tasks, such as those in the RULER benchmark. 
\end{itemize}

\begin{table*}[h!]
\centering
\begin{adjustbox}{width=0.8\columnwidth}
\begin{tabular}{c|ccc|cccc}
\specialrule{2.5pt}{1pt}{1pt}
\multirow{2}{*}{\textbf{Sequence Length}} & \multicolumn{3}{c}{\textbf{LoRA Method (Rank=32)}} & \multicolumn{4}{c}{\textbf{LoRA Rank (Method=\ourlora)}} \\
& LoRA & LoRA+ & \ourlora & 8 & 16 & 32 & 64 \\
\specialrule{1.5pt}{1pt}{1pt}
2048 & 10.77 & 10.98 & 10.99 & 11.00 & 11.00 & 10.99 & 10.99 \\
\hline
\cellcolor[gray]{.9} 8192 & 10.29 & 10.49 & 10.45 & 10.45 & 10.46 & 10.45 & 10.44 \\
\hline
16384 & 10.91 & 11.18 & 11.14 & 11.03 & 11.10 & 11.14 & 11.14 \\
\hline
32768 & 12.34 & 12.66 & 12.64 & 12.37 & 12.53 & 12.64 & 12.64 \\
\hline
65536 & 13.99 & 14.36 & 14.26 & 13.84 & 14.08 & 14.26 & 14.28 \\
\hline
\end{tabular}
\end{adjustbox}
\caption{\textbf{Mamba-2-Hybrid LoRA Ablations.} We fine-tune Mamba-2-Hybrid with a context size of 8192 with \ourattnsp using different LoRA variants. We consider LoRA, LoRA+, and \ourlorasp (ours) and evaluate perplexity on the PG-19 dataset. We observe that all LoRA variants yield similar perplexity results. However, as depicted in \cref{fig:mamba_ruler_avg_hylora} and \cref{fig:lora_rank_comp}, different LoRA variants yield substantially different performances on more complex tasks.}
\label{table:mamba_lora_ablations}
\end{table*}

\section{Applying SE-Attention to Transformers}\label{sec:se_attn_on_transformers}

In this section, we show that \ourattnsp can also be applied to Transformer models and use Llama 1 7B \cite{touvron2023llama} as our representative Transformer model. When fine-tuning Llama 7B models, we found that using a fixed chunk size of $M=4096$ gave better downstream performance. All of our Llama models are fine-tuned with a context size of 16384. For \ourattn, we use a block size of 32, and a top-$\topk$ of 8. For \swattn, we use a window size of 4096, and \ssattnsp uses the default parameters in \cite{chen2023longlora}. Due to computational constraints, we do not consider fine-tuning with \fullattn. All Attention variants use the same RoPE \cite{su2024roformer} scaling as in~\cite{chen2023longlora}.

\begin{table*}[h!]
\centering
\begin{adjustbox}{max width=\textwidth}
\begin{tabular}{c|cccc|ccccccc|cccc}
\specialrule{2.5pt}{1pt}{1pt}
\multirow{2}{*}{\textbf{Attention}} & \multicolumn{4}{c}{\textbf{Eval Context Size (PG-19 PPL $\downarrow$)}} & \multicolumn{7}{c}{\textbf{Short Context Tasks ($\uparrow$)}} & \multicolumn{4}{c}{\textbf{Long Context Tasks ($\uparrow$)}} \\
& 2048 & 8192 & \cellcolor[gray]{.85} 16384 & 32768 & ARC-E & ARC-C & Hella. & LAMB. & PIQA & WG & Avg. & SWDE & SQA & SNQA & Avg. \\
\specialrule{1pt}{1pt}{1pt}
Non-fine-tuned & 8.63 & 17.62 & 105.78 & 244.32 & 73.19 & 41.38 & 66.64 & 54.38 & 76.28 & 62.51 & \cellcolor[gray]{.9} 62.40 & 38.52 & 20.75 & 12.36 & \cellcolor[gray]{.9} 23.88 \\
\specialrule{1pt}{1pt}{1pt}
\swattn & 8.80 & 8.17 & 8.35 & 10.32 & 74.20 & 43.09 & 75.00 & 71.96 & 77.42 & 67.48 & \cellcolor[gray]{.9} 68.19 & 82.81 & 24.41 & 21.06 & \cellcolor[gray]{.9} 42.76 \\
\ssattn & 9.32 & 8.64 & 8.58 & 10.42 & 74.20 & 42.58 & 73.84 & 69.94 & 77.80 & 67.72 & \cellcolor[gray]{.9} 67.68 & 80.56 & 23.36 & 19.65 & \cellcolor[gray]{.9} 41.19 \\
\specialrule{1pt}{1pt}{1pt}
\ourattnsp & 8.76 & 8.13 & 8.01 & 9.28 & 75.04 & 44.37 & 75.02 & 71.03 & 78.02 & 67.17 & \cellcolor[gray]{.9} 68.44 & 84.79 & 24.59 & 21.36 & \cellcolor[gray]{.9} 43.58 \\
\bottomrule
\end{tabular}
\end{adjustbox}
\caption{\textbf{Fine-tuning Llama1 with \ourattnsp outperforms fine-tuning with \ssattnsp and \swattnsp on natural language tasks.} We fine-tune Llama1 with a context size of 16384 using various Attention variants. Similar to applying \ourattnsp to Mamba-2-Hybrid, here we again observe that fine-tuning Llama with \ourattnsp improves upon \swattnsp and \ssattn.}
\label{table:llama_ppl_and_lm_harness_sp}
\end{table*}

\noindent \textbf{Perplexity.} Fine-tuning Llama with \ourattnsp leads to better generalization on context sizes smaller and larger than the one used for fine-tuning. In \cref{table:llama_ppl_and_lm_harness_sp}, we observe that fine-tuning with \ourattnsp preserves the performance of the non-fine-tuned model at smaller context sizes, and offers greater generalization to larger context sizes,
as measured on PG-19 validation perplexity.

\noindent \textbf{LM Harness.} Fine-tuning Llama with \ourattnsp yields a stronger performance on long-context tasks on the LM Evaluation Harness benchmark. As shown in \cref{table:llama_ppl_and_lm_harness_sp}, fine-tuning with \swattn, \ssattn,  and \ourattnsp all improve upon the non-fine-tuned model on shorter context tasks and perform similarly. However, \ourattnsp gives a greater performance on long context tasks.

\begin{figure*}[t]
\vskip 0.2in
\begin{center}
\centerline{\includegraphics[width=0.6\textwidth]{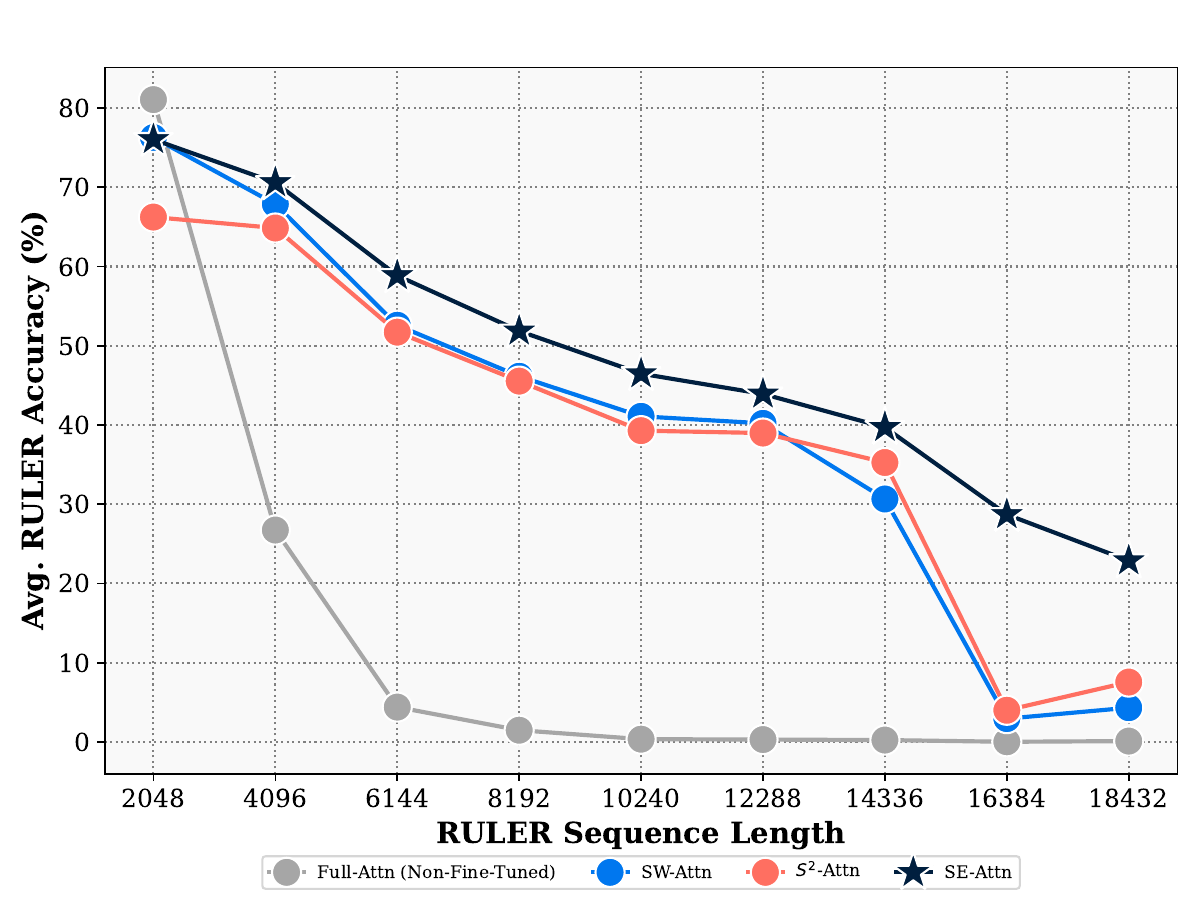}}
\caption{\textbf{Fine-tuning with \ourattnsp outperforms \swattnsp and \ssattnsp on the RULER benchmark when applied to Llama1.} We fine-tune Llama1 with a context size of 16384 using various Attention variants. We average over eleven RULER tasks, as explained in \cref{sec:ruler_aggregation}. Fine-tuning with \ourattnsp consistently outperforms \swattnsp and \ssattnsp even when evaluating on context sizes beyond the fine-tuning size.}
\label{fig:llama_ruler_avg}
\end{center}
\vskip -0.2in
\end{figure*}

\noindent \textbf{RULER.} Fine-tuning Llama with \ourattnsp produces a model with stronger performance on RULER tasks than fine-tuning with \swattnsp and \ssattn,  as illustrated in \cref{fig:llama_ruler_avg}. On average, \swattnsp and \ssattnsp perform similarly, however, fine-tuning with \ourattnsp improves performance by $\sim$5\%. Despite all models having a similar PPL up to a context size of 16k as shown in \cref{table:llama_ppl_and_lm_harness_sp}, we observe a substantial difference between the RULER performance of models fine-tuned with our \ourattnsp and those fine-tuned with the \swattnsp and \ssattn, again suggesting that PPL is not the most accurate assessment of long-context performance, as discussed in \cref{sec:additional_lora_ablations}.

\section{Beyond Mamba-2-Hybrid}\label{sec:zamba2}
In this section, we apply \ourattnsp to the Zamba2 1.2B model \cite{glorioso2024zamba2}. Zamba2 uses shared Attention blocks with different LoRA adapters for the shared blocks. Despite its intricate architecture, we can replace these layers with \ourattnsp and fine-tune it to perform on larger context sizes. The pre-trained model was trained with a context size of 4096 and we are able to efficiently fine-tune it with a context size of 12,288. We benchmark our model on NIAH tasks from RULER and provide results in \cref{table:zamba_comparisons}. We compare to \cite{yang2024gated} and \cite{yang2024parallelizing_no_conv} and demonstrate competitive performance to SOTA models.

\begin{table}[h!]
\centering
\resizebox{\columnwidth}{!}{
\begin{tabular}{ll|cccc|cccc|cccc}
\toprule
& & \multicolumn{4}{c|}{S-NIAH-1} & \multicolumn{4}{c|}{S-NIAH-2} & \multicolumn{3}{c}{S-NIAH-3} \\
\cmidrule{3-14}
Model & & 1K & 2K & 4K & 8K & 1K & 2K & 4K & 8K & 1K & 2K & 4K & 8K \\
\midrule
DeltaNet & & 97.4 & 96.8 & 99.0 & 98.8 & 98.4 & 45.6 & 18.6 & 14.4 & 85.2 & 47.0 & 22.4 & - \\
Mamba2 & & 99.2 & 98.8 & 65.4 & 30.4 & 99.4 & 98.8 & 56.2 & 17.0 & 64.4 & 47.6 & 4.6 & - \\
Gated DeltaNet & & 98.4 & 88.4 & 91.4 & 91.8 & 100.0 & 99.8 & 92.2 & 29.6 & 86.6 & 84.2 & 27.6 & - \\
\midrule
Zamba2-Hybrid & & 100.0 & 99.2 & 99.4 & 0.0 & 100.0 & 100.0 & 75.0 & 0.0 & 94.4 & 75.4 & 37.6 & 0.0 \\
\textbf{Zamba2-Hybrid + SE-Attn} & & 100.0 & 100.0 & 100.0 & 41.0 & 100.0 & 100.0 & 100.0 & 42.0 & 100.0 & 100.0 & 94.0 & 34.0 \\
\bottomrule
\end{tabular}
}
\caption{\textbf{\ourattnsp can efficiently extend the context length of Zamba2-Hybrid}. We fine-tune Zamab2-Hybrid using \ourattnsp and evaluate on RULER NIAH tasks.}
\label{table:zamba_comparisons}
\vspace{-9mm}
\end{table}

\section{Additional Fine-Tuning Results on Mamba-2-Hybrid}\label{sec:additional_ft_results}
\begin{figure*}[t]
\vskip 0.2in
\begin{center}
\centerline{\includegraphics[width=0.95\textwidth]{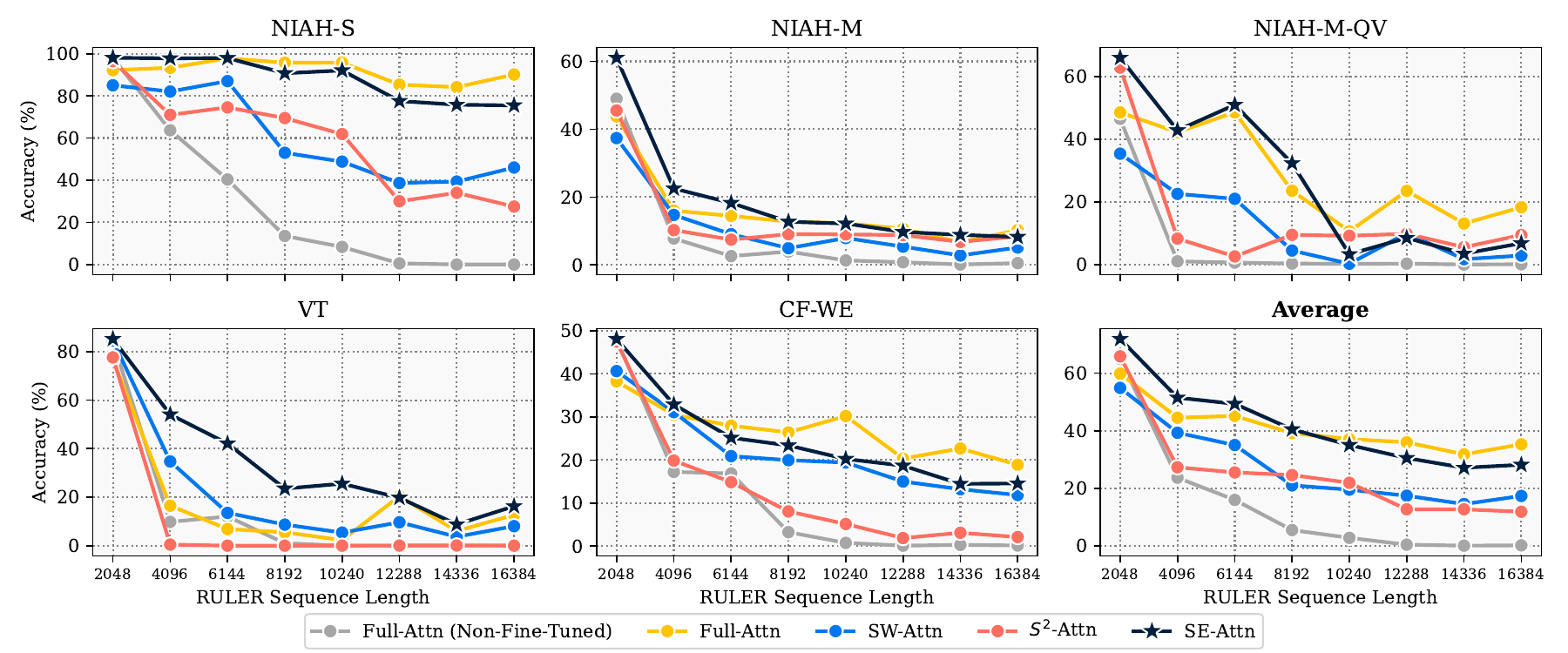}}
\caption{\textbf{Mamba-2-Hybrid RULER benchmark fine-tuned on natural language.} We fine-tune Mamba-2-Hybrid with a context size of 8192 using multiple Attention variants and evaluate on the RULER benchmark (see \cref{sec:ruler_aggregation} for definitions of each metric). On average, fine-tuning with \ourattnsp yields a performance close to the paragon model fine-tuned with \fullattnsp. We observe the biggest gain in needle-in-a-haystack (NIAH) tasks, as well as variable tracking (VT), all of which require strong recall capabilities enabled by our \ourattn.}
\label{fig:mamba_ruler_sp}
\end{center}
\vskip -0.2in
\end{figure*}

In this section, we provide additional benchmarks on the Mamba-2-Hybrid model. 
We begin by expanding upon \cref{fig:mamba_ruler_avg_hylora}(a) with additional RULER tasks in \cref{fig:mamba_ruler_sp}. Here, we observe that Mamba-2-Hybrid fine-tuned with \ourattnsp consistently outperforms \swattnsp and \ssattnsp on all RULER tasks across a broad range of context sizes---even beyond the 8192 fine-tuning context size.

\subsection{Fine-Tuning on Natural Language + Code}
\begin{figure*}[ht]
\vskip 0.2in
\begin{center}
\centerline{\includegraphics[width=0.95\textwidth]{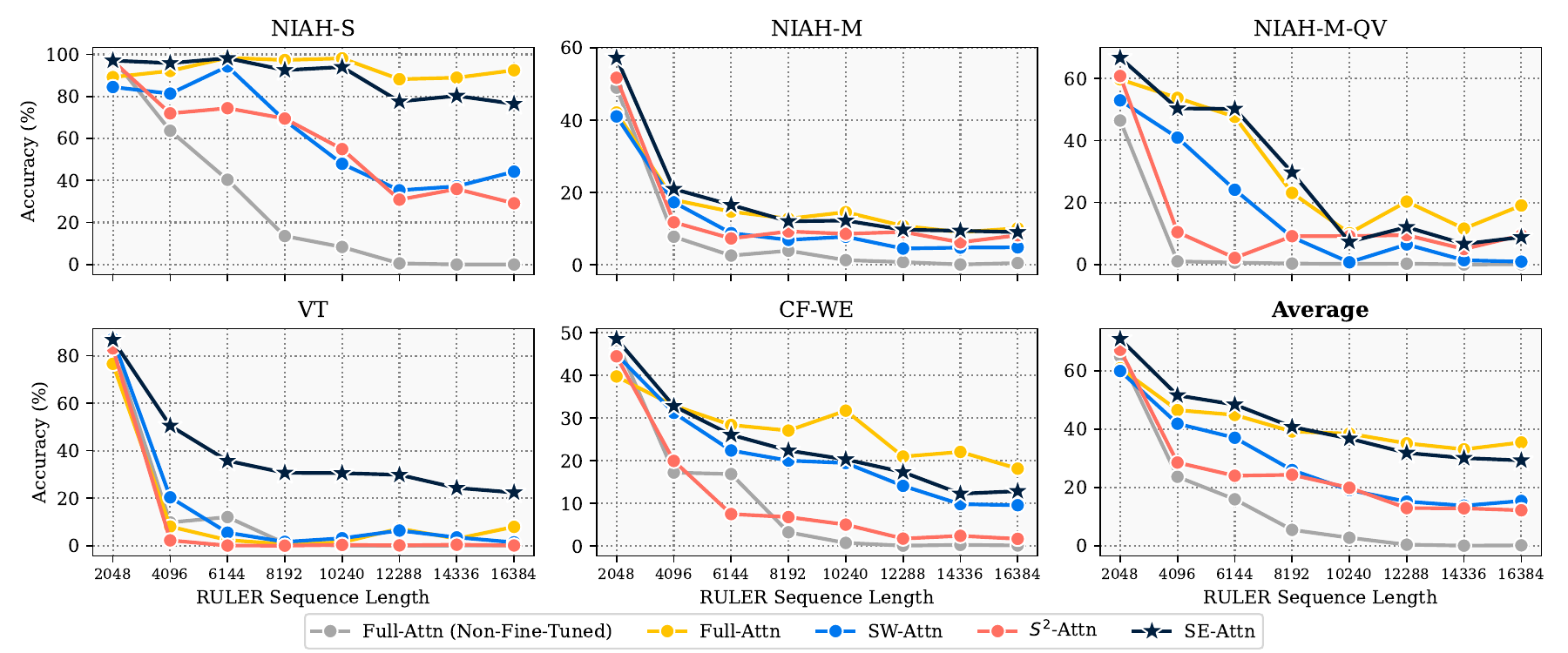}}
\caption{\textbf{Mamba-2-Hybrid RULER benchmark fine-tuned on natural language + code.} We fine-tune Mamba-2-Hybrid with different Attention layers on a dataset that consists of 70\% natural language and 30\% code and evaluate on RULER. Compared to fine-tuning on only natural language (as in \cref{fig:mamba_ruler_sp}), we see a substantial improvement on tasks like variable tracking (VT), and needle-in-a-haystack tasks (NIAH), both of which require strong recall capabilities enabled by fine-tuning with \ourattn.}
\label{fig:mamba_ruler_locp}
\end{center}
\vskip -0.2in
\end{figure*}

Next, we consider fine-tuning on a dataset that augments natural language with a code dataset. In particular, we construct a dataset consisting of 70\% natural language, and 30\% C++ code extracted from the Lots of Code \cite{lotsofcode} dataset. We provide PG-19 validation PPL results, as well as results on tasks from the LM Evaluation Harness suite in \cref{table:mamba_ppl_and_lm_harness_locp}. Similar to fine-tuning on natural language only (as in \cref{table:mamba_ppl_and_lm_harness_sp}), we observe that fine-tuning with \ourattnsp yields the strongest performance on downstream tasks. We provide performance on RULER in \cref{fig:mamba_ruler_locp}. We again observe that fine-tuning with \ourattnsp yields the strongest perfromance compared to other efficient Attention layers. Moreover, compared to fine-tuning only on natural language, here we observe a greater improvement on the Variable Tracking task. As explained in \cref{sec:ruler_aggregation}, this task requires the model to keep track of the values of variables that are defined and overridden throughout the input context, and must then return the variables equal to some value. This requires strong recall capabilities, which fine-tuning with \ourattnsp enables, and is amplified by the use of code data in the fine-tuning data mix.

\begin{table*}[h!]
\centering
\begin{adjustbox}{max width=\textwidth}
\begin{tabular}{c|cccc|ccccccc|cccc}
\specialrule{2.5pt}{1pt}{1pt}
\multirow{2}{*}{\textbf{Attention}} & \multicolumn{4}{c}{\textbf{Eval Context Size (PG-19 PPL $\downarrow$)}} & \multicolumn{7}{c}{\textbf{Short Context Tasks ($\uparrow$)}} & \multicolumn{4}{c}{\textbf{Long Context Tasks ($\uparrow$)}} \\
& 2048 & \cellcolor[gray]{.85} 8192 & 16384 & 32768 & ARC-E & ARC-C & Hella. & LAMB. & PIQA & WG & Avg. & SWDE & SQA & SNQA & Avg. \\
\specialrule{1pt}{1pt}{1pt}
 Non-fine-tuned & 10.72 & 14.99 & 19.35 & 26.37 & 69.91 & 37.97 & 67.62 & 69.84 & 76.06 & 65.04 & \cellcolor[gray]{.9} 64.41 & 85.60 & 15.18 & 3.65 & \cellcolor[gray]{.9} 34.81 \\
\specialrule{1pt}{1pt}{1pt}
\fullattn & 10.94 & 10.23 & 10.36 & 11.09 & 70.12 & 38.14 & 67.40 & 69.34 & 75.08 & 64.56 & \cellcolor[gray]{.9} 64.11 & 84.88 & 25.99 & 19.61 & \cellcolor[gray]{.9} 43.49 \\
\hline
\swattn & 10.94 & 10.76 & 11.81 & 13.43 & 69.70 & 38.82 & 67.51 & 69.38 & 75.41 & 64.88 & \cellcolor[gray]{.9} 64.28 & 84.52 & 24.44 & 15.30 & \cellcolor[gray]{.9} 41.42 \\
\hline
\ssattn & 10.86 & 12.89 & 14.67 & 16.36 & 69.95 & 37.88 & 67.45 & 69.94 & 76.01 & 64.72 & \cellcolor[gray]{.9} 64.32 & 86.50 & 16.65 & 8.61 & \cellcolor[gray]{.9} 37.25 \\
\specialrule{1.5pt}{1pt}{1pt}
\ourattn & 10.95 & 10.41 & 11.07 & 12.50 & 70.45 & 38.91 & 67.39 & 69.07 & 75.08 & 64.96 & \cellcolor[gray]{.9} 64.31 & 85.24 & 26.14 & 18.08 & \cellcolor[gray]{.9} 43.15 \\
\hline
\end{tabular}
\end{adjustbox}
\caption{\textbf{Fine-tuning Mamba-2-Hybrid with \ourattnsp on a natural language + code dataset outperforms fine-tuning with \ssattnsp and \swattnsp on natural language tasks.} We fine-tune Mamba-2-Hybrid with a context size of 8192 using various Attention variants on a dataset that consists of 70\% natural language and 30\% code.  We evaluate PG-19 validation perplexity (PPL) and observe that fine-tuning with \ourattnsp yields better perplexity scores than \ssattnsp and \swattn. On short-context tasks from the LM Harness suite, all models perform similarly. On long context tasks from the LM Harness suite, \ourattnsp outperforms \ssattnsp and \swattn.}
\label{table:mamba_ppl_and_lm_harness_locp}
\end{table*}

\subsection{Fine-Tuning on PG-19}
\begin{figure*}[ht]
\vskip 0.2in
\begin{center}
\centerline{\includegraphics[width=0.95\textwidth]{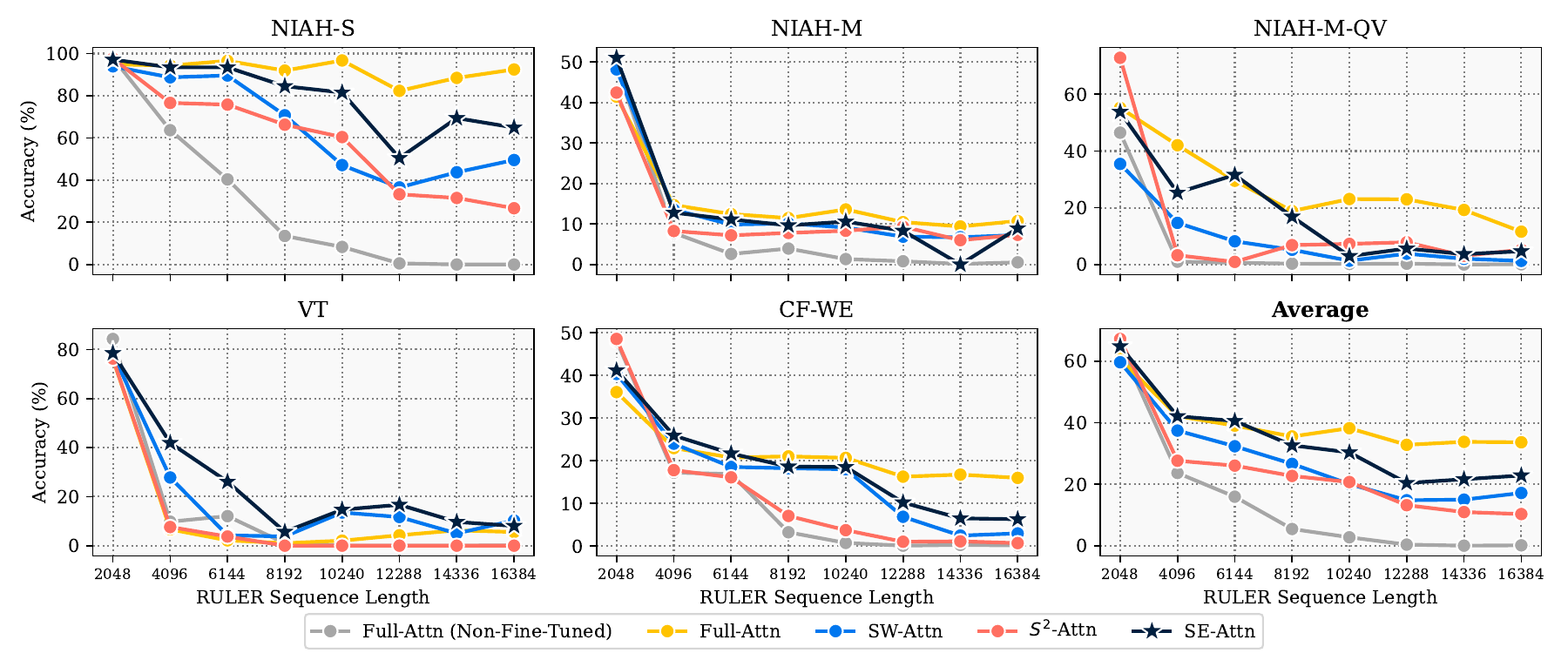}}
\caption{\textbf{Mamba-2-Hybrid RULER benchmark fine-tuned on PG-19.} We fine-tune Mamba-2-Hybrid with different Attention layers on the PG-19 \cite{rae2019compressive} dataset and then evaluate on RULER. Compared to fine-tuning on other natural language datasets with a greater variety of text as in \cref{fig:mamba_ruler_sp}, here we observe a degradation in performance across all models, likely due to a distribution shift in the PG-19 data and the RULER tasks.}
\label{fig:mamba_ruler_pg19}
\end{center}
\vskip -0.2in
\end{figure*}

Next, we consider fine-tuning on the PG-19 \cite{rae2019compressive} dataset. We provide perplexity results on the validation set, along with LM Harness task results in \cref{table:mamba_ppl_and_lm_harness_pg19}. Compared to fine-tuning on a different natural language dataset, and a natural language + code dataset as in \cref{table:mamba_ppl_and_lm_harness_sp} and \cref{table:mamba_ppl_and_lm_harness_locp}, here we obtain lower perplexity scores due to the lack of distribution shift. Interestingly, compared to fine-tuning on the previously mentioned datasets, we observe a slight degradation on LM Harness tasks. Moreover, in \cref{fig:mamba_ruler_pg19} we plot RULER performance when fine-tuning on PG-19, and here we also see a decay in performance compared to fine-tuning on the previous datasets. This suggests that the PG-19 dataset may be too far out of distribution for these long-context retrieval task. Nevertheless, we see that fine-tuning with \ourattnsp yields the best performance across all of these tasks.

\begin{table*}[h!]
\centering
\begin{adjustbox}{max width=\textwidth}
\begin{tabular}{c|cccc|ccccccc|cccc}
\specialrule{2.5pt}{1pt}{1pt}
\multirow{2}{*}{\textbf{Attention}} & \multicolumn{4}{c}{\textbf{Eval Context Size (PG-19 PPL $\downarrow$)}} & \multicolumn{7}{c}{\textbf{Short Context Tasks ($\uparrow$)}} & \multicolumn{4}{c}{\textbf{Long Context Tasks ($\uparrow$)}} \\
& 2048 & \cellcolor[gray]{.85} 8192 & 16384 & 32768 & ARC-E & ARC-C & Hella. & LAMB. & PIQA & WG & Avg. & SWDE & SQA & SNQA & Avg. \\
\specialrule{1pt}{1pt}{1pt}
 Non-fine-tuned & 10.72 & 14.99 & 19.35 & 26.37 & 69.91 & 37.97 & 67.62 & 69.84 & 76.06 & 65.04 & \cellcolor[gray]{.9} 64.41 & 85.60 & 15.18 & 3.65 & \cellcolor[gray]{.9} 34.81 \\
\specialrule{1pt}{1pt}{1pt}
\fullattn & 10.73 & 10.04 & 10.14 & 10.91 & 67.34 & 37.12 & 66.80 & 68.62 & 74.32 & 62.67 & \cellcolor[gray]{.9} 62.81 & 84.88 & 25.35 & 18.46 & \cellcolor[gray]{.9} 42.90 \\
\hline
\swattn & 10.72 & 10.59 & 11.64 & 13.25 & 66.96 & 37.54 & 66.83 & 68.79 & 74.54 & 63.30 & \cellcolor[gray]{.9} 62.99 & 84.79 & 22.54 & 14.09 & \cellcolor[gray]{.9} 40.48 \\
\hline
\ssattn & 10.78 & 12.74 & 14.42 & 16.11 & 69.36 & 38.14 & 67.45 & 69.90 & 76.12 & 64.72 & \cellcolor[gray]{.9} 64.28 & 86.50 & 17.00 & 7.74 & \cellcolor[gray]{.9} 37.08 \\
\specialrule{1.5pt}{1pt}{1pt}
\ourattn & 10.73 & 10.22 & 10.84 & 12.28 & 67.42 & 37.88 & 66.48 & 69.22 & 73.88 & 61.88 & \cellcolor[gray]{.9} 62.80 & 85.15 & 23.06 & 16.46 & \cellcolor[gray]{.9} 41.55 \\
\hline
\end{tabular}
\end{adjustbox}
\caption{\textbf{Fine-tuning Mamba-2-Hybrid with \ourattnsp on PG-19 outperforms fine-tuning with \ssattnsp and \swattnsp on long-context natural language tasks.} We fine-tune Mamba-2-Hybrid with a context size of 8192 using various Attention variants on PG-19.  We evaluate PG-19 validation perplexity (PPL) and observe that fine-tuning with \ourattnsp yields better perplexity scores than \ssattnsp and \swattn. On short-context tasks from the LM Harness suite, \ssattnsp has the strongest performance. On long context tasks from the LM Harness suite, \ourattnsp outperforms \ssattnsp and \swattn.}
\label{table:mamba_ppl_and_lm_harness_pg19}
\end{table*}

\subsection{Additional Long-context Benchmarks}\label{sec:appendix_lctx_tasks}
In this section, we evaluate our Mamba-2-Hybrid model fine-tuned on only language data on additional long-context tasks. First, we test our models' long-context understanding on the LongBench benchmark and report results in \cref{table:mamba_longbench}. Here, we observe that our fine-tuned models improve over the pre-trained baseline models, suggesting that these models' long-context understanding has improved. Moreover, we see that models fine-tuned with our \ourattnsp match the performance of the model fine-tuned with \fullattnsp.

We previously considered RULER as a means of assessing our models' in-context recall capabilities. While RULER can assess our models' performance on synthetic in-context recall tasks, we also want to evaluate our their performance on ``real-world'' in-context tasks. As such, in \cref{table:mamba_rw_icr}, we evaluate on the tasks considered in \cite{arora2024just}. We observe that fine-tuning with our \ourattnsp is able to match the performance of fine-tuning with \fullattnsp on these tasks.

\begin{table*}[t]
\centering
\begin{adjustbox}{max width=0.99\textwidth}
\begin{tabular}{c|ccc|ccc|ccc|cc|cc|c}
\specialrule{2.5pt}{1pt}{1pt}
\multirow{2}{*}{\textbf{Attention}} & \multicolumn{3}{c}{\textbf{Single-Doc QA ($\uparrow$)}} & \multicolumn{3}{c}{\textbf{Multi-Doc QA ($\uparrow$)}} & \multicolumn{3}{c}{\textbf{Summarization ($\uparrow$)}} & \multicolumn{2}{c}{\textbf{Few-shot ($\uparrow$)}} & \multicolumn{2}{c}{\textbf{Code ($\uparrow$)}} &  \\
& NQA & QQA & MFQ & HQA & 2WM & Mus & GvR & QMS & MNs & TRC & TQA & LCC & RBP & Avg. \\
\specialrule{1pt}{1pt}{1pt}
 Non-fine-tuned & 0.92 & 3.78 & 13.92 & 4.72 & 8.26 & 1.54 & 9.65 & 3.71 & 23.23 & 59.0 & 63.22 & 60.55 & 31.99 & 21.88 \\ 
\specialrule{1pt}{1pt}{1pt}
\fullattn & 3.25 & 7.15 & 16.35 & 8.03 & 9.51 & 3.69 & 25.4 & 17.64 & 26.05 & 64.5 & 82.09 & 56.67 & 55.58 & 28.92 \\ 
\specialrule{1pt}{1pt}{1pt}
\swattn & 2.29 & 7.52 & 17.07 & 7.72 & 9.4 & 3.41 & 18.44 & 11.87 & 23.94 & 64.5 & 81.48 & 58.33 & 53.06 & 27.62 \\ 
\ssattn & 1.39 & 5.58 & 16.43 & 5.31 & 7.88 & 1.83 & 13.64 & 5.64 & 26.86 & 57.0 & 72.97 & 56.62 & 36.87 & 23.69 \\ 
\specialrule{1pt}{1pt}{1pt}
\ourattn & 3.19 & 9.37 & 16.0 & 8.31 & 9.31 & 3.94 & 24.65 & 14.39 & 26.86 & 66.5 & 80.82 & 58.71 & 53.65 & 28.90 \\ 
\bottomrule
\end{tabular}
\end{adjustbox}
\caption{\textbf{\ourattnsp matches the performance of \fullattnsp on long-context understanding tasks}. We fine-tune Mamba-2-Hybrid models using various attention mechanisms on 14 LongBench \cite{bai2023longbench} tasks.}
\label{table:mamba_longbench}
\vspace{-0.5cm}
\end{table*}

\begin{table}[h]
    \centering
    \resizebox{0.7\columnwidth}{!}{%
    \begin{tabular}{c|cccccc|c}
    \specialrule{2.5pt}{1pt}{1pt}
    \textbf{Attention} &   SWDE &   SQUADv2 &   FDA &   TriviaQA &   NQ &   Drop &   Avg \\
    \specialrule{1pt}{1pt}{1pt}
     Non-fine-tuned     &  85.60 &     50.11 & 76.41 &      22.60 &      6.62 &   3.20 & 40.76 \\
     \specialrule{1pt}{1pt}{1pt}
     \fullattn    &  85.24 &     50.09 & 80.22 &      25.75 &      6.79 &   3.55 & 41.94 \\
     \specialrule{1pt}{1pt}{1pt}
     \swattn     &  84.61 &     50.10 & 78.31 &      25.51 &      6.76 &   3.62 & 41.48 \\
     \ssattn   &  86.41 &     50.11 & 81.49 &      23.63 &      6.84 &   3.32 & 41.97 \\
     \specialrule{1pt}{1pt}{1pt}
     \ourattn &  85.96 &     50.09 & 82.12 &      26.39 &      7.17 &   3.76 & 42.58 \\
    \bottomrule
    \end{tabular}
    }
    \caption{\textbf{\ourattnsp matches the performance of \fullattnsp on real-world in-context tasks}. We fine-tune Mamba-2-Hybrid models with different attention layers. We do not truncate inputs to 2K tokens when evaluating as was done in \cite{arora2024just}.}
    \label{table:mamba_rw_icr}
\end{table}

\subsection{Evaluating with Efficient Attention}
\begin{figure*}[h]
    \centering
    \subfigure[]{\includegraphics[width=0.425\textwidth]{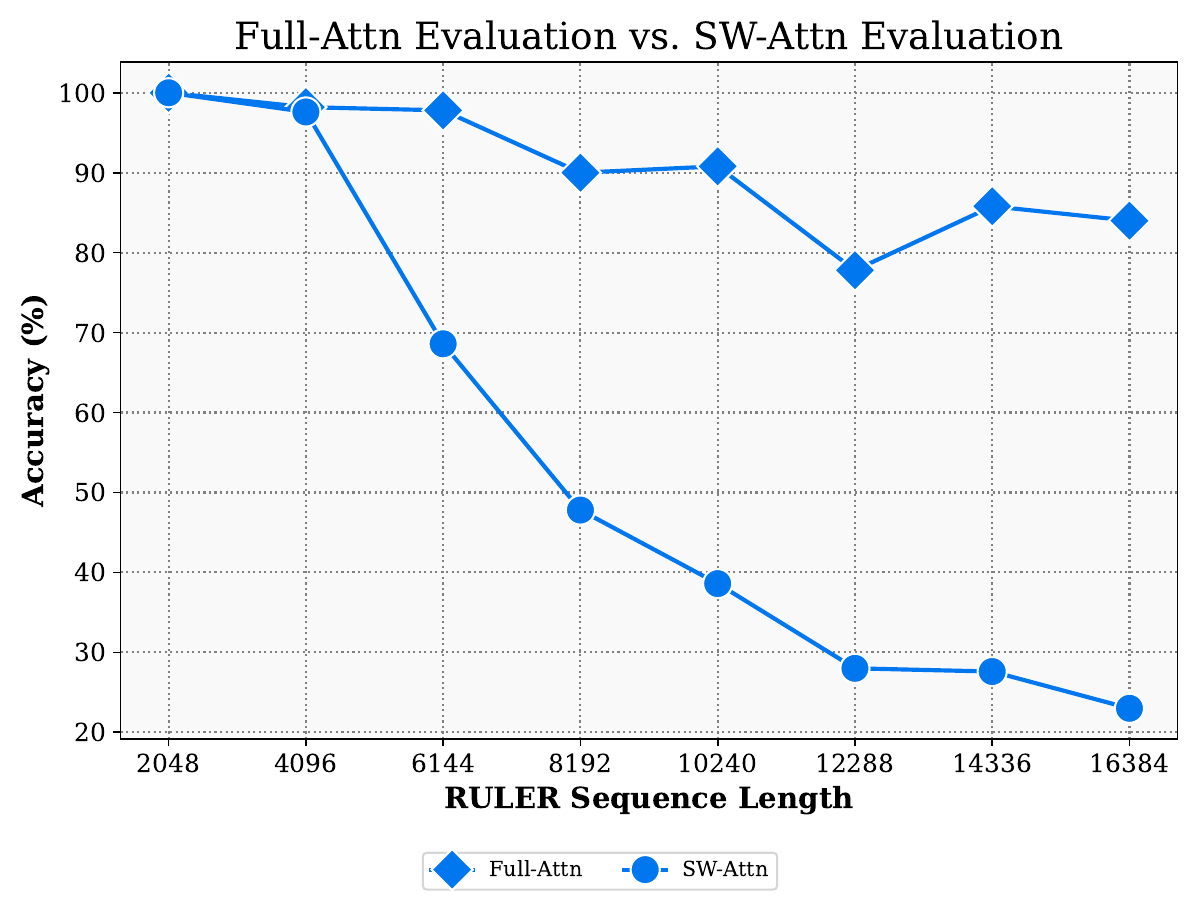}}
    \hfill
    \subfigure[]{\includegraphics[width=0.425\textwidth]{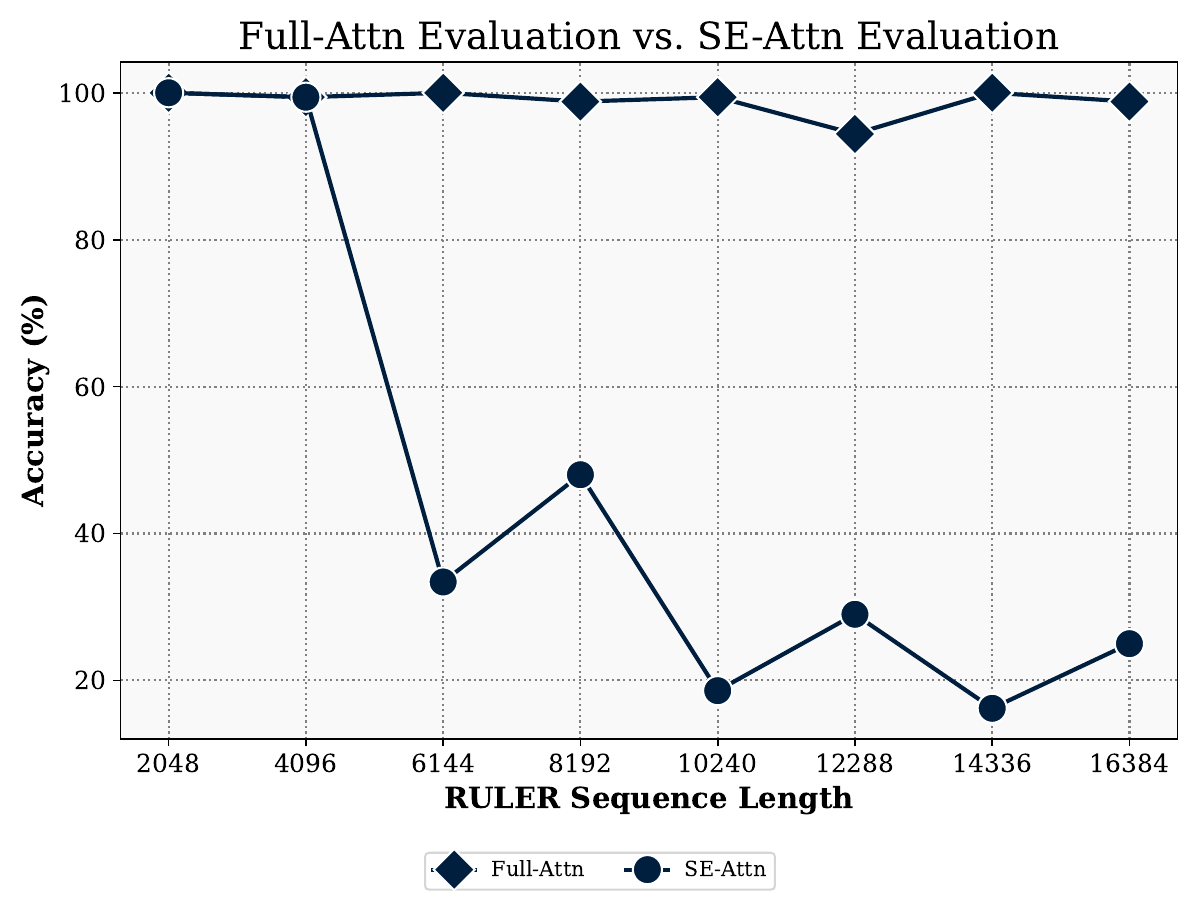}}
    \caption{\textbf{Evaluating with efficient Attention mechanisms on RULER does not do as well as evaluating with standard full Attention.} We fine-tune Mamba-2-Hybrid with \swattnsp (a) and \ourattnsp (b) and then evaluate on the NIAH-Single-1 RULER task using either the same Attention layer used for fine-tuning, or \fullattn. We observe that fine-tuning with efficient Attention layers (\ourattnsp and \swattn) and then using \fullattnsp during evaluation yields better results. Lines with a circle marker denote models fine-tuned with an efficient Attention mechanism, and then evaluated with the same Attention mechanism; lines with a diamond marker were fine-tuned with an efficient Attention mechanism, but evaluated with \fullattn. We omit results for evaluating models fine-tuned with \ssattnsp as this Attention mechanism does not support arbitrary sequence lengths during inference.}
    \label{fig:ft_attn_niah}
    \vspace{-0.2cm}
\end{figure*}

In \cref{sec:eval_with_full}, we explained how we fine-tune our models using an efficient Attention mechanism, and then evaluate with full attention. In \cref{table:ft_hydra_eval_hydra}, we showed that evaluating our models with efficient Attention mechanisms produced strong results on perplexity benchmarks. However, we did not observe similarly positive results on more complex tasks, such as those in RULER. As illustrated in \cref{fig:ft_attn_niah}, using the same efficient Attention layer used during fine-tuning does not perform as well as evaluating with \fullattnsp (we omit results for models fine-tuned with \ssattnsp and evaluated with \ssattnsp as the implementation of \ssattnsp does not support evaluating with it at arbitrary sequence lengths). Hence, our training pipeline involves training a model with an efficient Attention layer, and reverting to \fullattnsp during inference. Moreover, we again note that perplexity is a misleading metric in regard to measuring a model's performance on long context tasks. Though we observed strong performance on perplexity when evaluating with these efficient Attention layers, this did not translate to stronger performance on tasks that require a strong recall capability, such as those in RULER. 

\section{Retrieval with Landmark Tokens}\label{sec:landmark_ablation}
Landmark Attention \cite{mohtashami2023landmark} was recently introduced as a way for Transformer models to process long sequences by inserting ``landmark'' tokens into the sequence whose representations would then be used as summaries of the blocks of tokens that came before them. At a high level, our approach in \ourattnsp is similar. However, we simplify the process of compressing blocks of tokens by forgoing the use of landmark tokens and instead using Attention to summarize them, as described in \cref{sec:hydra_retrieval}. Moreover, to learn which blocks to retrieve, we do not rely on a complex Grouped Softmax function, and instead use a simple Cross-Attention score to ascertain relevance. In this way, we implement retrieval natively into the model's architecture. 

\begin{figure}[h!]
\vskip 0.2in
\begin{center}
\centerline{\includegraphics[width=0.6\columnwidth]{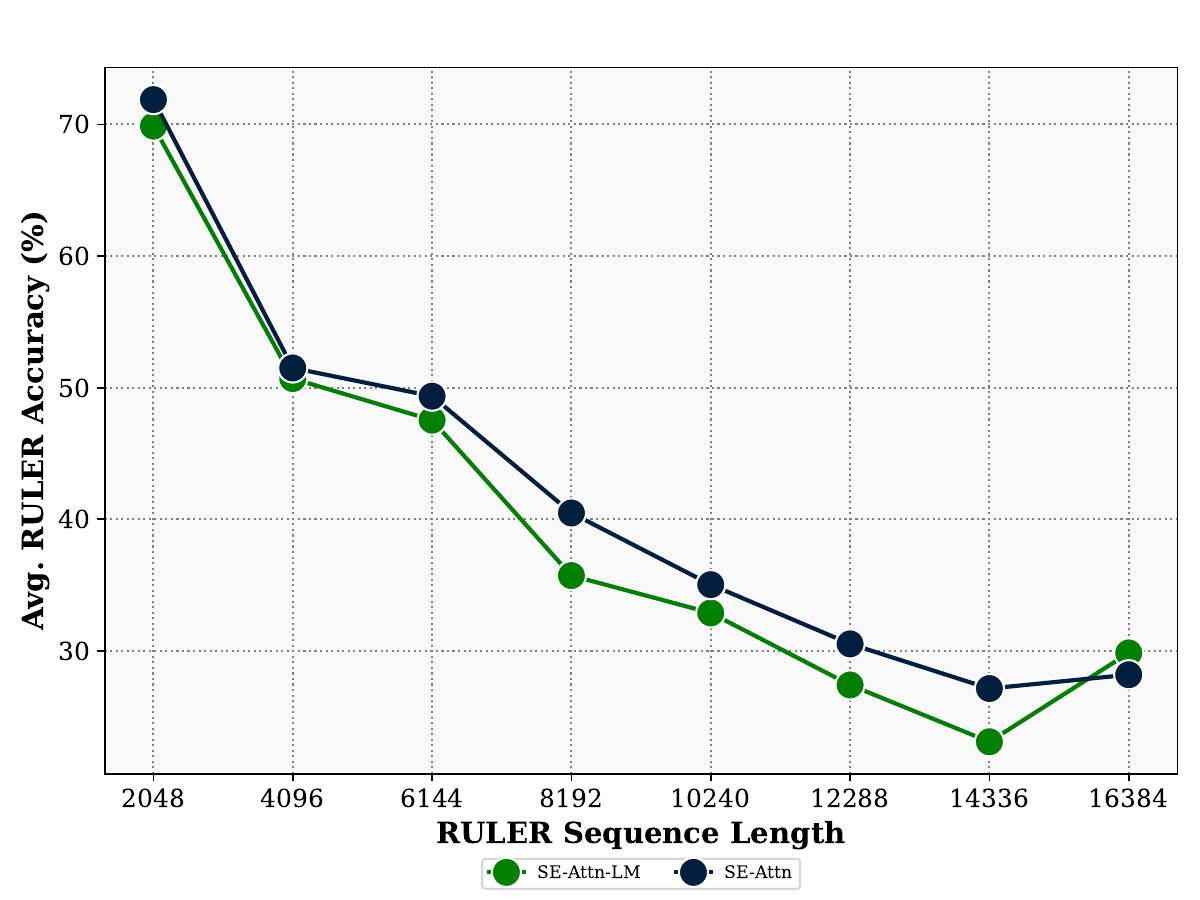}}
\caption{\textbf{Summarizing memory blocks via average pooling of attention yields stronger performance than summarizing them using ``landmark'' tokens.} We fine-tune Mamba-2-Hybrid using our \ourattn, and \ourattn-LM, which summarizes memory blocks with a landmark token (similar to \cite{mohtashami2023landmark}) instead of using the average of the Self Attention output of memory blocks, as in \ourattn. We find that our simpler \ourattnsp produces a stronger model, likely due to the easier training task, which does not require adapting the model to leverage landmark tokens for compression.}
\label{fig:mp_vs_lm}
\end{center}
\vskip -0.2in
\end{figure}

In this section, we consider a variant of \ourattn, which we refer to as \ourattn-LM, that is inspired by Landmark Attention. Namely, instead of using our Attention-based compression to construct summaries of memory blocks, we insert a non-learnable ``landmark'' token into each memory block, and use the cross-attention between this token and the memory tokens as the summary of the memory block. We compare the performance of this variant to our summaries computed using average pooling of Attention scores (see \cref{sec:hydra_retrieval}) in \cref{fig:mp_vs_lm}. Here, we see that \ourattnsp yields better performance. We suspect this is because using a non-learnable landmark token to summarize memory blocks is too challenging of a task to accomplish using standard LoRA. While full fine-tuning (without LoRA) may improve the performance of \ourattn-LM, this is beyond the scope of our work as we prioritize efficiency.

\section{Training Details}\label{sec:training_details}
\begin{figure}[ht]
\vskip 0.2in
\begin{center}
\centerline{\includegraphics[width=0.6\columnwidth]{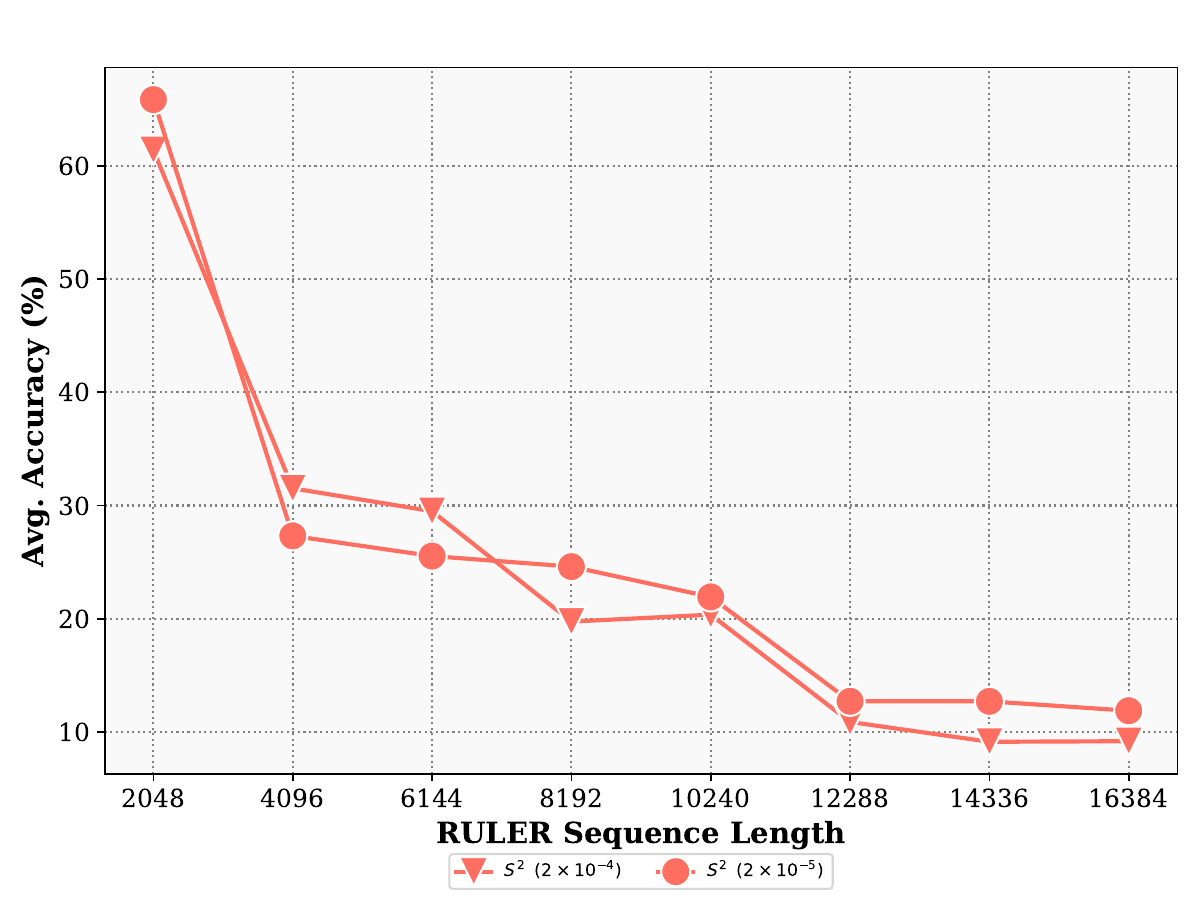}}
\caption{\textbf{A larger learning rate does not improve the performance of $S^2$.} Here we fine-tune a Mamba-2-Hybrid model using \ssattnsp with two different learning rates: $2\times 10^{-4}$ and $2\times 10^{-5}$. The learning rate used in \cite{chen2023longlora} is $2\times 10^{-5}$, which we found to work well. For all other Attention layers, we found $2\times 10^{-4}$ to offer a slight improvement over $2\times 10^{-5}$.}
\label{fig:s2_lr_comparison}
\end{center}
\vskip -0.2in
\end{figure}

Our fine-tuning recipe largely follows \cite{chen2023longlora}, with the exception of using a larger learning rate for \ourattn, \swattn, and \fullattnsp models ($2\times 10^{-4}$ vs. the default $2\times 10^{-5}$ for $S^2$). We found that using a larger learning rate for $S^2$ did not improve its performance, as shown in \cref{fig:s2_lr_comparison}, so we used the default $2\times 10^{-5}$ learning rate for all $S^2$ fine-tuning experiments.

All of our experiments are conducted on a single 8xA100 node. We fine-tune for 1000 steps on a total of 0.5B tokens. We fine-tune Mamba-2-H with a context size of 8192 with 8 accumulation steps and a per-device batch size of 1. We fine-tune Llama1 with with a context size of 16384 tokens with 4 accumulation steps and a per-device batch size of 1. We use FlashAttention-2 \cite{dao2023flashattention} and DeepSpeed Stage 2 \cite{rasley2020deepspeed}.

\section{Runtime and Memory Analysis}\label{sec:theoretical_runtime}
The runtime of \ourattnsp is lower than \fullattnsp and \ssattnsp on large contexts. For an input with sequence length $L$, \ourattnsp first constructs $U=\frac{L}{S}$ memory blocks of size $S$. Attention is applied on each. Thus, the complexity of this operation is $O(\mdim U S^2) = O(\mdim LS)$. Next, the input sequence is split into $T=\frac{L}{M}$ chunks of size $M$. Cross-Attention is then applied between each chunk's query and each memory block's compressed representation; the cost of this is $O(\mdim MTU) = O(\mdim \frac{L^2}{S})$. Cross-Attention is then applied between each chunk's query tokens and the concatenations of the chunk's key tokens and memory tokens. Since \ourattnsp retrieves $K$ blocks with $S$ tokens in each, the cost of this Cross-Attention for each chunk is $O(\mdim M(SK+M)) = O(\mdim M^2)$ since $SK < M$. The cost for all chunks is therefore $O(\mdim TM^2) = O(\mdim LM)$. The total cost of \ourattnsp is therefore $O(\mdim LS + \mdim LM + \mdim \frac{L^2}{S})$. For sufficiently large $S$, the runtime of \ourattnsp is faster than \fullattnsp and \ssattn, especially on large contexts, and is similar to that \swattn.

\section{RULER Task Definitions}
The RULER benchmark \cite{hsieh2024ruler} consists of four different task categories: retrieval, multi-hop tracing, aggregation, and question answering. In this paper, we focus only on the retrieval, multi-hop tracing, and aggregation tasks (we provide question answering results on the LM Evaluation Harness benchmark). These three categories span eleven different tasks as explained below.

\subsection{Needle-in-a-Haystack (NIAH) Tasks}
RULER consists of 8 different NIAH tasks. These tasks embed ``needles'' in a string of noise. These needles are typically key-value pairs, and the goal is to return the value of a key. These tasks are characterized by six parameters:

\begin{itemize}
\item{
    type\_haystack (TH): This specifies the type of noise to embed the key in. The choices are ``repeat'' which constructs noise as in \cite{mohtashami2023landmark}, ``essay'' which will use sentences from the Paul Graham essays \cite{paulgrahamessay}, or ``needle'' in which case each sentence will define a new key-value pair.
}
\item{
    type\_needle\_k (TK): This specifies the type of the needle's key. The options are ``words'', in which case the key is a word (in the form of adjective-noun, e.g., spiritual-oven), or ``uuids'' in which case the key is a UUID. 
}
\item{
    type\_needle\_v (TV): This specifies the type of the needle's value. It can either be ``numbers'' in which case the value is a 7-digit number, or it can be ``uuids'' in which case the value is a UUID.
}
\item{
    num\_needle\_k (NK): This specifies the number of key-value pairs to embed in the haystack.
}
\item{
    num\_needle\_v (NV): This specifies how many different values a key is assigned. If greater than 1, the goal is output all the values of they key.
}
\item{
    num\_needle\_q (NQ): This specifies the number of different keys the model must return the value for.
}
\end{itemize}

\begin{table}[h]
\centering
\resizebox{0.6\columnwidth}{!}{%
\begin{tabular}{c|c|c|c|c|c|c}
\specialrule{2.5pt}{1pt}{1pt}
NIAH Task & TH & TK & TV & NK & NV & NQ \\
\hline
Single 1 & repeat & words & numbers & 1 & 1 & 1 \\
\hline
Single 2 & essay & words & numbers & 1 & 1 & 1 \\
\hline
Single 3 & essay & words & uuids & 1 & 1 & 1 \\
\hline
Multikey 1 & essay & words & numbers & 4 & 1 & 1 \\
\hline
Multikey 2 & needle & words & numbers & 1 & 1 & 1 \\
\hline
Multikey 3 & needle & uuids & uuids & 1 & 1 & 1 \\
\hline
Multivalue & essay & words & numbers & 1 & 4 & 1 \\
\hline
Multiquery & essay & words & numbers & 1 & 1 & 4 \\
\hline
\end{tabular}
}
\caption{\textbf{RULER NIAH definitions.} The ``Needle-in-a-Haystack'' (NIAH) tasks in the RULER benchmarks are defined by 6 parameters which modulate the difficulty of the tasks. We consider 8 different NIAH tasks as defined above (these are the default NIAH tasks in the RULER library).}
\end{table}

\subsection{Multi-hop Tracing Tasks}
RULER considers a ``variable tracking'' task that is a form of coreference resolution. In this task, a sequence of variables are defined throughout noisy text as in \cite{mohtashami2023landmark}. New variables are defined as previous ones, and a final value is assigned to a particular variable. The goal is to be able to trace back which variables have also been assigned the final value, i.e., determine which variables refer to the final value. We use the default num\_chains=1 and num\_hops=4 parameters

\subsection{Aggregation Tasks}
RULER considers two aggregation tasks, common words extraction (CWE), and frequent words extraction (FWE). In CWE, the context consists of list of words, and the goal is to return the most common words. We use the default parameters freq\_cw=30, freq\_ucw=3, and num\_cw=10. In FWE, the context consists of random word strings, and the goal is to return the ones that appear the most frequently. We use the default alpha=2 parameter for this.

\subsection{Aggregating RULER Tasks}\label{sec:ruler_aggregation}
We aggregate the eleven RULER tasks above into six groups:
\begin{enumerate}
\item \textit{NIAH-S}: NIAH Single 1, NIAH Single 2, NIAH Single 3.
\item \textit{NIAH-M}: NIAH Multikey 1, NIAH Multikey 2, NIAH Multikey 3.
\item \textit{NIAH-M-QV}: NIAH Multivalue, NIAH Multiquery.
\item \textit{VT}: Variable Tracking.
\item \textit{CF-WE}: Common Words Extraction (CWE) and Frequent Words Extraction (FWE).
\item \textit{Average}: The average of all eleven tasks above.
\end{enumerate}

\section{Task Abbreviations}\label{sec:task_abbrev}
\textbf{LM Harness}: Hella=HellaSwag, LAMB=LAMBADA, WG=WinoGrande, SQA=ScrollsQAsper, SNQA=ScorllsNarrativeQA.

\noindent \textbf{LongBench}: NQA=NarrativeQA, QQA=QasperQA, MFQ=MultiFieldQA, HQA=HotpotQA, \\ 2WM=2WikiMultihopQA, Mus=Musiqe, GvR=GovReport, QMS=QMSum, MNs=MultiNews, TRC=TREC, TQA=TriviaQA, SSM=SamSum, RBP=RepoBench-P.